\RequirePackage{fix-cm}
\documentclass{article}

\usepackage{microtype}
\usepackage{graphicx}
\usepackage{subcaption}
\usepackage{booktabs}
\usepackage{multirow}
\usepackage[table]{xcolor}
\usepackage{hyperref}
\definecolor{mygreen}{RGB}{0, 150, 0}
\definecolor{myred}{RGB}{200, 0, 0}
\newcommand{\inc}[1]{{\color{mygreen}\footnotesize ~(+#1)}}
\newcommand{\dec}[1]{{\color{myred}\footnotesize ~(-#1)}}

\newcommand{\AppTOCTitle}{%
  \begin{center}
    {\scshape Overview}
  \end{center}
  \noindent This material provides supplementary details to the main paper,
  including the following sections:\par
  \vspace{0.35em}
}
\newcommand{\apptocleader}{%
  \nobreak\leaders\hbox{\kern.18em.\kern.18em}\hfill\nobreak
}
\newcommand{\apptocmark}[1]{%
  \textup{(}#1\textup{)}%
}
\newcommand{\apptocA}[2]{%
  \par\noindent
  \begingroup
  \def\apptoclabel{#1}%
  \hangindent=1.05em\hangafter=1
  \textbullet\hspace{0.35em}%
  \apptocAparse#2\apptocnil
  \apptocleader
  \hbox to 1.5em{\hfil\hyperref[#1]{\textbf{\pageref{#1}}}}%
  \par\endgroup
}
\newcommand{\apptocB}[2]{%
  \par\noindent
  \begingroup
  \def\apptoclabel{#1}%
  \hangindent=2.15em\hangafter=1
  \hspace*{1.1em}--\hspace{0.35em}%
  \apptocBparse#2\apptocnil
  \apptocleader
  \hbox to 1.5em{\hfil\hyperref[#1]{\pageref{#1}}}%
  \par\endgroup
}
\def\apptocAparse#1 \quad#2\apptocnil{%
  \hyperref[\apptoclabel]{\textbf{\apptocmark{#1} #2}}%
}
\def\apptocBparse#1 \quad#2\apptocnil{%
  \hyperref[\apptoclabel]{\apptocmark{#1} #2}%
}

\usepackage{tabularx}
\usepackage{amsmath}
\usepackage[utf8]{inputenc} 
\usepackage{tcolorbox}
\usepackage{enumitem}
\usepackage{listings}
\usepackage{float}
\usepackage{fancyvrb}
\usepackage{amssymb}

\usepackage[accepted]{icml2026}

\makeatletter
\renewcommand{\printAffiliationsAndNotice}[1]{\global\icml@noticeprintedtrue%
  {\let\thefootnote\relax\footnotetext{\hspace*{-\dimexpr\footnotesep+\parindent\relax}\raggedright%
      $^*$Equal contribution. $^\dagger$Corresponding author.\par
      \Notice@String
    }
  }
}
\makeatother

\usepackage{mathtools}
\usepackage{amsthm}

\newcommand{\mypara}[1]{\smallskip\noindent\textbf{#1}}

\usepackage[capitalize,noabbrev]{cleveref}

\theoremstyle{plain}

\theoremstyle{definition}

\theoremstyle{remark}

\icmltitlerunning{Benchmarking and Evolving Reason-Reflect-Rectify for Reflective Visual Generation}

\begin{document}

\twocolumn[
  \icmltitle{Benchmarking and Evolving \texorpdfstring{\\}{ } Reason-Reflect-Rectify for Reflective Visual Generation}

  \begin{center}
    {\bfseries
    Junjie Wang$^{1,*}$ \quad
    Xinghua Lou$^{2,3,*}$ \quad
    Jason Li$^{4}$ \quad
    Ye Tian$^{5}$ \quad
    Keyu Chen$^{1}$ \quad
    Yulin Li$^{1}$\\
    Bin Kang$^{6}$ \quad
    Jacky Mai$^{7}$ \quad
    Yanwei Li$^{8}$ \quad
    Zhuotao Tian$^{1,3,\dagger}$ \quad
    Liqiang Nie$^{1,3}$
    }

    \vspace{0.35em}
    {\fontsize{9.5pt}{11pt}\selectfont
    \textsuperscript{1}Harbin Institute of Technology, Shenzhen \quad
    \textsuperscript{2}University of Science and Technology of China \quad
    \textsuperscript{3}Shenzhen Loop Area Institute\\
    \textsuperscript{4}Nanyang Technological University \quad
    \textsuperscript{5}Peking University \quad
    \textsuperscript{6}University of Chinese Academy of Sciences\\
    \textsuperscript{7}Hong Kong Baptist University \quad
    \textsuperscript{8}Shanghai Jiao Tong University
    }
  \end{center}

  \hypersetup{pdfauthor={Junjie Wang, Xinghua Lou, Jason Li, Ye Tian, Keyu Chen, Yulin Li, Bin Kang, Jacky Mai, Yanwei Li, Zhuotao Tian, Liqiang Nie}}

  \icmlkeywords{Multimodal, Visual Generation, Large Language Models}
    
  \vskip 0.3in

]

\printAffiliationsAndNotice{\icmlEqualContribution}

\begin{abstract}
Text-to-Image (T2I) models and Unified Multimodal Models (UMMs) have achieved remarkable progress in visual generation. 
However, their reliance on a single-pass generation paradigm limits their ability to handle complex prompts requiring iterative refinement. To enable multi-round Reflective Visual Generation (RVG), we formalize the \emph{Reason--Reflect--Rectify} (R$^3$) loop as a core framework and introduce R$^3$-Bench, a benchmark of over 600 expert-annotated instances that quantifies iterative reasoning and rectification capabilities. 
Evaluation on R$^3$-Bench reveals a critical gap: while state-of-the-art models can identify generation errors, they fail to generate actionable rectification instructions. 
To bridge this gap, we propose {R$^3$-Refiner}, a dual-stage framework leveraging Group Relative Policy Optimization (GRPO) and a Hierarchical Reward Mechanism (HRM) to better align rectification with reflective reasoning.
Experiments show that {R$^3$-Refiner} achieves significant improvements on R$^3$-Bench (+12.0\% in Reflective Verdict Score, +9.0\% in Rectification Score), 
and can be seamlessly integrated with various MLLMs to enhance the generation quality of different T2I models on GenEval++ and T2I-CompBench. 
Code is available at \href{https://github.com/xiaomoguhz/R3-Bench}{https://github.com/xiaomoguhz/R3-Bench}.
\end{abstract}

\begin{figure}[!t]
\centering
\includegraphics[width=\columnwidth]{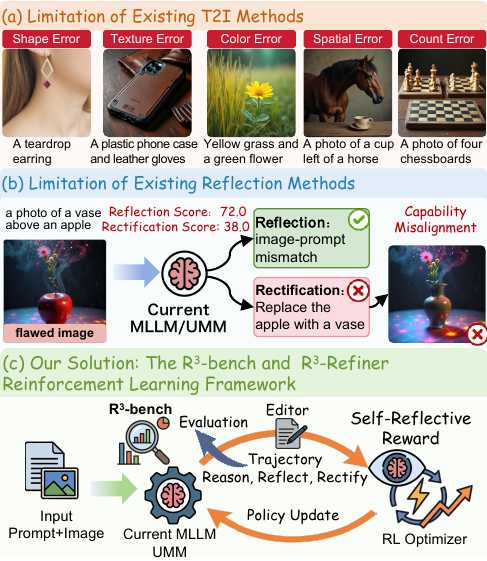}
\caption{\textbf{(a)} Existing text-to-image (T2I) models often struggle with compositional prompts, resulting in diverse visual generation errors. \textbf{(b)} Reflective visual generation aims to mitigate these errors, yet current models suffer from a capability misalignment: they accurately diagnose flaws (strong reasoning) but fail to execute valid corrections (weak rectification). \textbf{(c)} To bridge this gap, we introduce R$^3$-Bench and R$^3$-Refiner, a reinforcement learning framework that leverages the model's strong reasoning as a self-reflective reward to optimize rectification policies.}
\label{fig:teaser}
\end{figure}

\section{Introduction}
\label{sec:introduction}

Text-to-Image (T2I) task~\cite{sd3, sdxl, flux, flux1kontext, ldm} has achieved remarkable success with diffusion models. 
Building upon these advancements, Unified Multimodal Models (UMMs)~\cite{showo2, mogao, janus_pro, lumina_dimoo, mmada, illume+, emu3_5} integrate the reasoning capabilities of Multimodal Large Language Models (MLLMs)~\cite{blip2, kimi, keyevl1_5, seedvl1_5, deepseekvl, llava, minigpt4, internvl3_5} to further enhance visual generation capabilities. 
However, as illustrated in Fig.~\ref{fig:teaser}(a), these models still struggle with compositional prompts because the open-loop, single-pass generation paradigm lacks mechanisms for error rectification. 
To overcome this limitation, a transition to a multi-round Reflective Visual Generation (RVG) paradigm is necessary.

\begin{figure*}[t]
\centering
\includegraphics[width=0.95\linewidth]{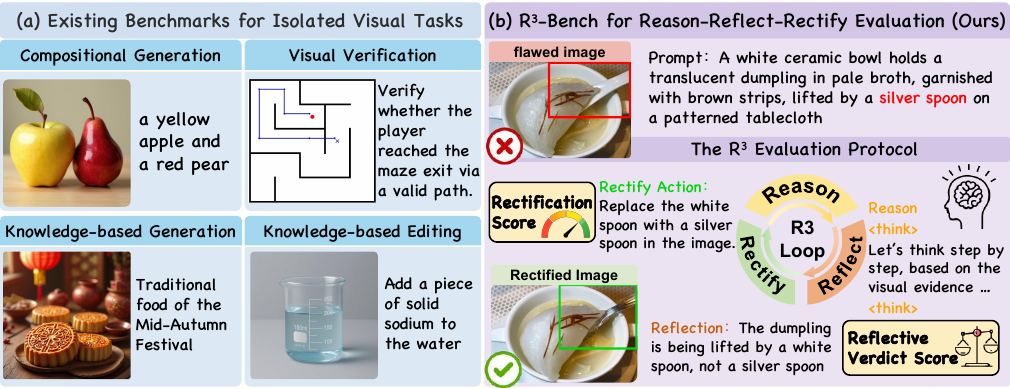}
\caption{\textbf{Comparison between existing benchmarks and R$^3$-Bench.} \textbf{(a)} Existing benchmarks predominantly evaluate image generation, editing, and visual verification as isolated tasks. \textbf{(b)} In contrast, R$^3$-Bench centers on the ``Reason-Reflect-Rectify"  loop for Reflective Visual Generation (RVG). As illustrated by the ``silver spoon'' example, the model first employs reasoning to diagnose inconsistency by providing a verdict and a reflective explanation. The accuracy of this diagnosis is quantified by the Reflective Verdict Score. Subsequently, the model generates a rectification action to guide the refinement process. The efficacy of this correction is measured by the Rectification Score, which assesses relative visual improvement.}
\label{fig:teaser2}
\end{figure*}

\paragraph{Insufficient Evaluation for RVG. }
Following the success of self-reflection mechanisms in Large Language Models (LLMs)~\cite{react, reflexion, Cot_valve, mitigating, seal} and MLLMs~\cite{sherlock, critic, llava_critic_r1, self_correct, self_refine}, recent visual generation studies~\cite{IRG, mure, ThinkMorph} have started exploring closed-loop RVG paradigms. 
However, advancing research in this direction is hindered by a critical \textit{evaluation} gap. 
As illustrated in Fig.~\ref{fig:teaser2}(a), existing benchmarks predominantly measure isolated capabilities, including attribute alignment~\cite{geneval++, geneval, dpgbench}, reasoning and knowledge-based generation~\cite{wise, kris}, or multidimensional understanding and generation~\cite{mme_unify, oneig, realunify}. 
None of these benchmarks quantifies the iterative reasoning processes integral to RVG, \textit{i.e.}, diagnosing visual inconsistencies, reflecting upon corrective strategies, and rectifying generated outputs.

\paragraph{The Proposed R$^3$-Bench. }
To effectively assess RVG capabilities, we formalize the critical competencies into the Reason-Reflect-Rectify (R$^3$) loop and introduce the R$^3$-Bench, which comprises 670 expert-annotated correction tasks covering both synthetic and real-world scenarios. Each task provides a textual prompt paired with a flawed generated image, requiring the model to output a structured response, consisting of a \textit{verdict} answer, a \textit{reflective} explanation, and a \textit{rectification} action for evaluation. We employ a dual evaluation protocol to comprehensively assess the model (Fig.~\ref{fig:teaser2}(b)). Specifically, we evaluate the diagnostic accuracy of the verdict and reflective explanation while quantifying the efficacy of the rectification action based on the relative visual improvement of the flawed image.

\paragraph{Key Observations. }
Using our benchmark, we find that even state-of-the-art visual reasoning and generation models fall short in these challenging scenarios. As shown in Fig.~\ref{fig:teaser}(b), although leading MLLMs~\cite{qwenvl2_5, omniverifier} can identify inconsistencies between textual prompts and generated images, they often fail to yield actionable rectification instructions. 
Such a discrepancy limits the effectiveness of the closed-loop RVG pipeline required for high-quality visual generation. 
This observation raises a critical question: \textit{Can we harness the model's strong discriminative capability as a reward signal to enhance its rectification ability through self-evolution? }

\paragraph{Our Solution.}

To address the issue, we propose \textit{R$^3$-Refiner}, a reinforcement-learning-based refinement framework for RVG, as shown in Fig.~\ref{fig:teaser}(c).
R$^3$-Refiner explicitly adopts the Reason--Reflect--Rectify (R$^3$) loop and aligns rectification behavior with reflective reasoning through structured self-reward. Given a misaligned text-image pair, R$^3$-Refiner first produces a structured R$^3$ trajectory, consisting of (i) a reasoning process that diagnoses visual inconsistencies, (ii) a reflective verdict that determines error types and correction necessity, and (iii) a rectification instruction specifying actionable edits.

R$^3$-Refiner adopts a \textit{Hierarchical Reward Mechanism} (HRM) for optimization, which decomposes supervision across different stages of the R$^3$ loop, as illustrated in Fig.~\ref{fig:training_flow}.
Specifically, the reasoning and reflection stages are supervised using constructed R$^3$ data, ensuring accurate inconsistency diagnosis and reliable reflective judgments.
For rectification, R$^3$-Refiner adopts a closed-loop verification process where the rectification instruction is executed by an external image editor to generate a revised image, which is subsequently re-evaluated by the model to derive a self-reward signal based on visual improvement. Experiments on Qwen2.5-VL and Qwen3-VL~\cite{Qwen3-VL} demonstrate that R$^3$-Refiner enhances both reflective
verdict score and rectification score of the baseline. Moreover, R$^3$-Refiner can be seamlessly integrated with various MLLMs to enhance the generation quality of different T2I models, such as Bagel, OmniGen2, and GPT-Image.

In summary, the contributions of this work are as follows:

\begin{itemize}
\item We introduce R$^3$-Bench, a benchmark that operationalizes RVG through the Reason-Reflect-Rectify (R$^3$) loop and evaluates models across diagnosis, reflection, and rectification capabilities.
\item Utilizing R$^3$-Bench, we identify a critical capability misalignment between discriminative reasoning and rectification execution. To address this, we propose R$^3$-Refiner, a dual-stage framework that optimizes the complete R$^3$ loop using Group Relative Policy Optimization (GRPO) and the HRM.

\item Experiments demonstrate that R$^3$-Refiner significantly improves performance on R$^3$-Bench (+12.0\% reflection, +9.0\% rectification) and can be integrated with various MLLMs to enhance the generation quality of various T2I models.
\end{itemize}

\section{Reason-Reflect-Rectify for Reflective Visual Generation}
\label{sec:r3_loop}
This section introduces the task definition and benchmark construction for the Reason-Reflect-Rectify (R$^3$) framework. We first formalize the iterative R$^3$ loop tailored for RVG tasks in Sec.~\ref{subsec:task_formalization}. Next, Sec.~\ref{subsec:bench_construction} describes the composition and construction pipeline of the benchmark. Finally, Sec.~\ref{subsec:evaluation_protocol} outlines the corresponding evaluation protocol.

\subsection{Task Formalization}
\label{subsec:task_formalization}
We formalize RVG as an iterative Reason-Reflect-Rectify (R$^3$) loop that progressively refines image generation outputs. Unlike the conventional single-pass generation paradigm, the R$^3$ loop involves iterative processing by an MLLM or UMM, consisting of three stages: a global binary verification of image-text consistency (\textit{reason}), detailed localization of semantic discrepancies (\textit{reflect}), and formulation of precise corrective actions (\textit{rectify}).

\mypara{The Iterative R$^3$ Loop.} 
Formally, at refinement step $t$, the task aims to learn a policy $\pi_{\theta}$ that produces a structured response $R_t$ conditioned on the textual prompt $P$ and the current image $\mathbf{I}^{(t)}$. Adhering to the R$^3$ framework, the structured response $R_t$ is a tuple comprising a verification answer $v_t$, a discrepancy explanation $e_t$, and a rectification action $a_t$:
\begin{equation}
R_t = \pi_{\theta}(P, \mathbf{I}^{(t)}) = \langle v_t, e_t, a_t \rangle,
\end{equation}
where $v_t$ serves as a global binary indicator for image-text consistency, $e_t$ localizes and explains semantic discrepancies in detail, and $a_t$ specifies targeted editing instructions necessary to address the identified inconsistencies. Then, the structured response $R_t$ is used for iterative refinement.

Specifically, at iteration $t$, if the verification answer $v_t$ indicates image-text misalignment (denoted as \textit{False}), the explanation and rectification tuple $\langle e_t, a_t\rangle$ instructs an external generative editor $\Phi$. 
Consequently, the editor modifies $\mathbf{I}^{(t)}$ to yield an improved image $\mathbf{I}^{(t+1)}$:
\begin{equation}
\mathbf{I}^{(t+1)} = \Phi(\mathbf{I}^{(t)}, \langle e_t, a_t\rangle).
\end{equation}
This iterative refinement continues until the image-text consistency is confirmed, with $v_t$ becoming \textit{True}. 

\subsection{Benchmark for Evaluation}
\label{subsec:bench_construction}
In this section, we introduce R$^3$-Bench, which assesses a model's proficiency in verifying image-text consistency, localizing semantic discrepancies, and formulating precise rectification actions. Below, we describe the dataset composition and construction pipeline in detail.

\begin{figure}[t] 
\centering
\includegraphics[width=.8\columnwidth]{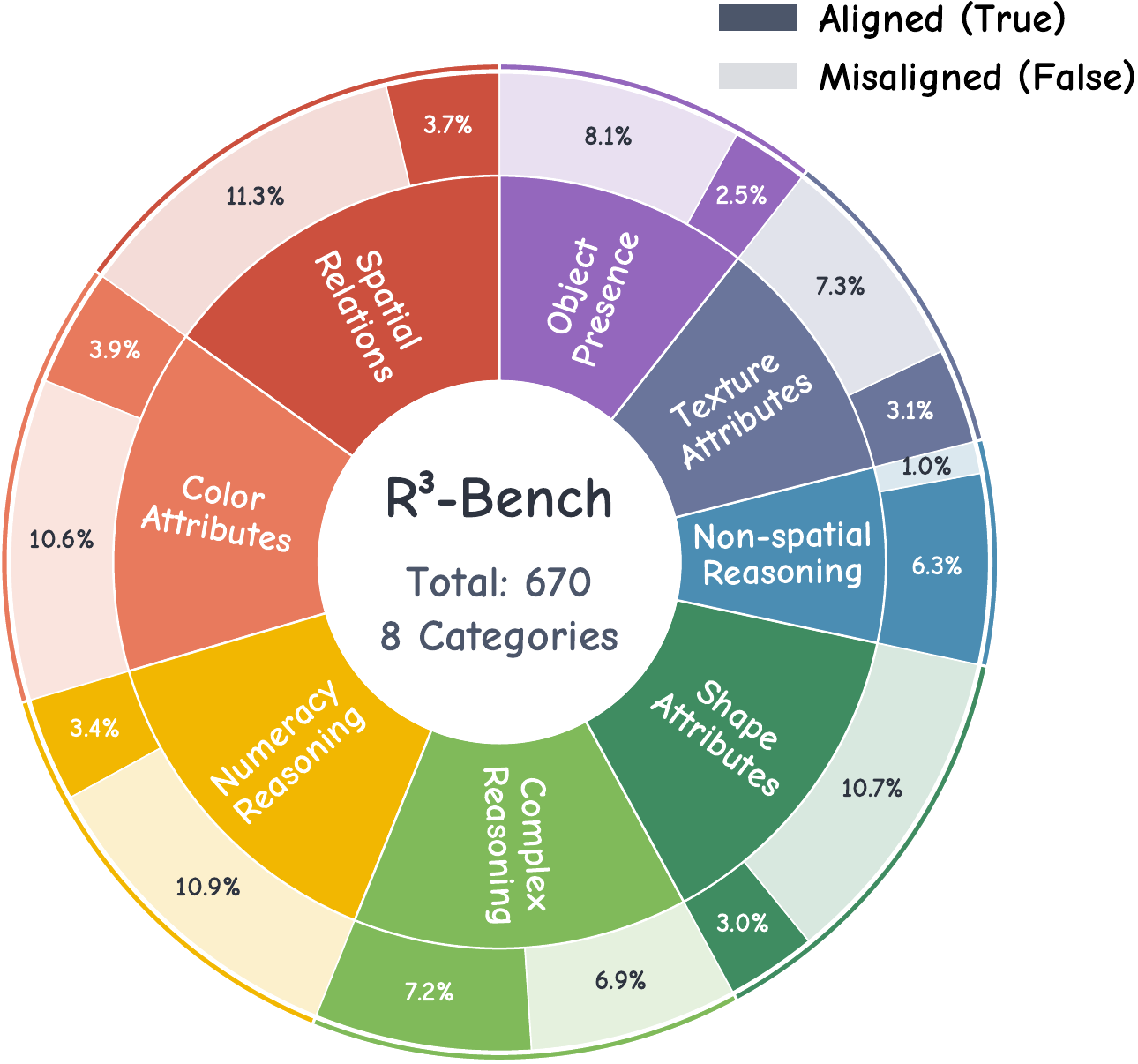}
\caption{\textbf{Overview of R$^3$-Bench.} The benchmark covers eight categories sourced from both real-world and model-generated data, comprising 222 aligned and 448 misaligned instances.}
\label{fig:r3_bench_overview}
\end{figure}

\mypara{Benchmark Overview.} As illustrated in Fig.~\ref{fig:r3_bench_overview}, R$^3$-Bench comprises 670 expert-annotated image-text pairs designed to evaluate the capabilities required for the iterative R$^3$ loop. 
R$^3$-Bench contains diverse error categories and balances aligned and misaligned instances to ensure comprehensive coverage. 
All instances are annotated with Ground-Truth (GT) verification labels to assess the accuracy of $v_t$. 
For misaligned instances, R$^3$-Bench provides additional GT explanations and related visual question-answering (VQA) tasks. These annotations are essential for benchmarking the full R$^3$ loop, enabling the evaluation of discrepancy reasoning $e_t$ and the effectiveness of rectification actions $a_t$ on rectified images.

\mypara{Benchmark Construction.}
We construct the benchmark through a multi-stage process designed to ensure clarity and diversity. To form an initial candidate pool, we aggregate data from complementary sources, combining error samples generated by T2I models~\cite{qwen_image} using prompts adapted from T2I-R1~\cite{t2i_r1} and GenEval++~\cite{geneval++} with image–text mismatches obtained by rewriting samples from the GEdit dataset~\cite{gedit} to incorporate diverse real-world domains.

This pool subsequently undergoes a cascaded filtering procedure (Sec.~\ref{subsec:data_pipeline}) to isolate preliminary matched and mismatched image–text pairs. Following this, we leverage MLLMs~\cite{Qwen3-VL} and LLMs~\cite{qwen3} to generate discrepancy explanations and VQA pairs that facilitate the evaluation of image improvements. To ensure the highest quality, human experts verify each case to refine annotations and eliminate ambiguous instances such as minor color distinctions or inconsistent scene atmospheres. This rigorous process yields 670 high-quality samples for balancing testing cost and data diversity. Examples and additional details are provided in Fig.~\ref{fig:r3_bench_examples} and Appendix~\ref{sec:test_set_details}.

\subsection{Evaluation Protocol}
\label{subsec:evaluation_protocol}
We evaluate models in the R$^3$ loop using a two-phase protocol. Let $\mathcal{S}=\{(P_i,\mathbf{I}_i^{(t)},v_i,e_i,Q_i)\}_{i=1}^N$ denote the test set, where $v_i$ and $e_i$ are the GT verification label and explanation, and $Q_i$ is a set of visual questions targeting key attributes in the prompt $P_i$.

\mypara{Phase I: Verdict-Reflection Alignment.}
This phase evaluates the accuracy of the model's generated verdict $\hat{v}_i$ and reflection $\hat{e}_i$. To quantify their combined accuracy, we introduce the Reflective Verdict Score ($\mathcal{S}_{\text{ref}}$), as detailed in Appendix~\ref{subsec:reflect_score}. Specifically, we compute a correctness score $s_i \in \{0, 1\}$ for each sample based on the GT $v_i$. For aligned samples where $v_i = \text{True}$, the score depends solely on the predicted verdict and is defined as $s_i = \mathbb{I}(\hat{v}_i = \text{True})$. Conversely, for misaligned samples where $v_i = \text{False}$, we enforce a stricter criterion that requires correctness in both the verdict and the reflection. Accordingly, we define $s_i = \mathbb{I}(\hat{v}_i = \text{False}) \cdot \mathcal{J}(e_i, \hat{e}_i)$, where the LLM-Judge $\mathcal{J}$~\cite{qwen3} returns 1 when the generated explanation $\hat{e}_i$ is semantically equivalent to $e_i$. Finally, $\mathcal{S}_{\text{ref}}$ is obtained by averaging $s_i$ over all samples.

\mypara{Phase II: Rectification Efficacy.}
This phase evaluates the effectiveness of the rectification action $\hat{a}_i$ generated by a model. The action is executed by an external image editor to produce a rectified image $\mathbf{I}_i^{(t+1)}$. Then, the improvement is assessed using a VQA-based alignment function $\mathcal{V}(\mathbf{I}, Q) \in [0, 1]$, which applies an external MLLM to answer the annotated question set $Q_i$ for both the original and rectified images (detailed in Appendix~\ref{subsec:eval_prompts}). We introduce the Rectification Score ($\mathcal{S}_{\text{rect}}$) to quantify the gain. Crucially, instead of absolute improvement, we calculate the normalized improvement relative to the initial error, computed over misaligned samples:
\begin{equation}
\mathcal{S}_{\text{rect}} = \frac{1}{N_{\texttt{neg}}} \sum_{i: v_i = \text{False}} \frac{\mathcal{V}(\mathbf{I}_i^{(t+1)}, Q_i) - \mathcal{V}(\mathbf{I}_i^{(t)}, Q_i)}{1 - \mathcal{V}(\mathbf{I}_i^{(t)}, Q_i)}.
\end{equation}
Here, $N_{\texttt{neg}}$ denotes the number of misaligned samples, and a higher $\mathcal{S}_{\text{rect}}$ indicates that a larger fraction of the initial discrepancy with respect to the target prompt is corrected.

\section{Method}
\label{sec:method}
The preceding section introduced R$^3$-Bench to evaluate model RVG capabilities. Evaluations on this benchmark (Tab.~\ref{tab:main_results}) reveal a critical misalignment in current MLLMs. Specifically, these models possess strong reasoning skills yet fail to translate them into effective image refinement. Motivated by this, we propose R$^3$-Refiner, a reinforcement learning framework that leverages these reasoning capabilities as feedback to better accomplish RVG tasks. R$^3$-Refiner can be seamlessly integrated with various MLLMs to enhance the generation quality of different T2I models.

\begin{figure*}[t]
\centering
\includegraphics[width=\linewidth]{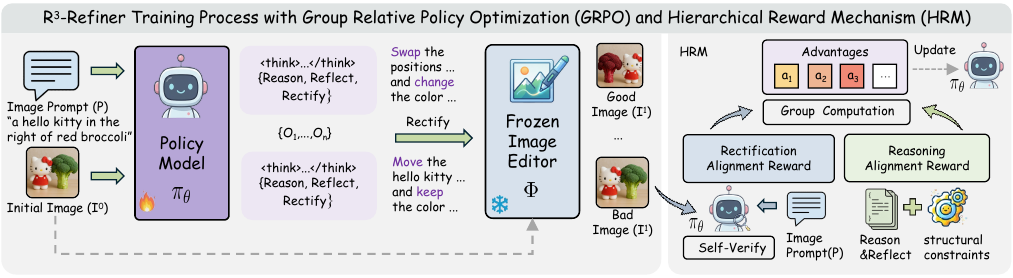}
\caption{ 
The policy $\pi_{\theta}$ samples $N$ structured trajectories via GRPO. 
The optimization is driven by a Hierarchical Reward Mechanism (HRM) comprising two stages: Reasoning Alignment ($R_\text{reason}$) and Rectification Alignment ($R_\text{rect}$).}
\label{fig:training_flow}
\end{figure*}

\subsection{R$^3$-Refiner}
\label{subsec:r3_refiner}

R$^3$-Refiner is a dual-stage framework that optimizes the complete R$^3$ loop using Group Relative Policy Optimization (GRPO)~\cite{deepseekmath}, as illustrated in Fig.~\ref{fig:training_flow}. 

Specifically, for each training input $\langle P, \mathbf{I}^{(0)} \rangle$, the policy $\pi_{\theta}$ samples a group of $N$ trajectories $\{o_1, \dots, o_N\}$. Consistent with the definitions in Sec.~\ref{subsec:task_formalization}, each trajectory is parsed as a tuple $o_i = \langle \hat{v}_i, \hat{e}_i, \hat{a}_i \rangle$, corresponding to the \textit{Reason}, \textit{Reflect}, and \textit{Rectify} components, respectively. We implement the dual-stage optimization of these trajectories via the following Hierarchical Reward Mechanism (HRM).

\mypara{Hierarchical Reward Mechanism.} To facilitate image-text consistency verification, explanation generation, and visual rectification within the R$^3$ loop, we propose the Hierarchical Reward Mechanism (HRM), which integrates two complementary optimization signals: (i) the Reasoning Alignment Reward ($R_\text{reason}$), which aligns verification with GT labels, and (ii) the Rectification Alignment Reward ($R_\text{rect}$), which assesses consistency between the rectified image and the original prompt via $\pi_{\theta}$ itself.
The design of HRM is motivated by our empirical observation that the model’s discriminative capability significantly exceeds its visual rectification capability (Tab.~\ref{tab:main_results}). We detail the formulation of these rewards below.

\begin{figure}[t] 
\centering
\includegraphics[width=0.9\columnwidth]{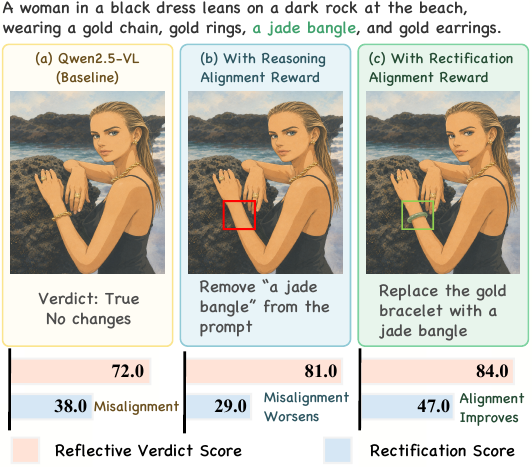}
\caption{\textbf{Effectiveness of HRM.} The Reasoning Alignment Reward improves verdict accuracy but induces \textit{Illusory Visual Rectification}. As shown in \textbf{(b)}, the policy learns to edit the prompt rather than refining the image. The Rectification Alignment Reward in \textbf{(c)} alleviates this behavior and encourages valid visual rectification.}
\label{fig:r3_bench}
\end{figure}

\mypara{Stage I: Reasoning Alignment Reward.}
This stage enhances the model's verification capabilities by targeting the \textit{Reason} phase (verdict $\hat{v}$). While the \textit{Reflect} phase (explanation $\hat{e}$) contains the reasoning chain, directly optimizing open-ended text generation via RL is unstable and computationally expensive. Following~\cite{omniverifier}, we instead posit that the verdict $\hat{v}$ serves as a reliable proxy for the quality of the underlying reasoning. Consequently, we design $R_{\text{reason}}$ to enforce the accuracy of the final verdict against the GT $v_{\text{gt}}$ (derived from our data construction  pipeline in Sec.~\ref{subsec:data_pipeline}):
\begin{equation}
    R_{\text{reason}} = \lambda_{\text{fmt}} \cdot \mathbb{I}(o_i \in \Omega) + \lambda_{\text{acc}} \cdot \mathbb{I}(\hat{v}_i = v_{\text{gt}}).
\end{equation}
Here, $\mathbb{I}(\cdot)$ denotes the indicator function. The coefficients $\lambda_{\text{fmt}}$ and $\lambda_{\text{acc}}$ weight the rewards for format compliance and prediction accuracy, respectively. The term $\Omega$ imposes the format reward that encourages each trajectory $o_i$ to follow the prescribed template provided in Appendix~\ref{subsec:training_prompt}. The accuracy term, $\mathbb{I}(\hat{v}_i = v_{\text{gt}})$, weighted by $\lambda_{\text{acc}}$, constitutes the principal optimization objective, guiding the model to ground its judgments explicitly on accurate visual evidence. By enforcing correctness in the final \textit{Reason} output ($\hat{v}$), we indirectly encourage logical consistency within the latent \textit{Reflect} explanation ($\hat{e}$).

\mypara{Illusory Visual Rectification.} Stage I improves verification by aligning the \textit{Reason} verdict with GT labels. A natural expectation is that stronger verification also leads to better rectification, because the policy should identify mismatches and then correct them. However, our empirical results, illustrated in Fig.~\ref{fig:r3_bench}(b), contradict this expectation. 
When trained exclusively with $R_{\text{reason}}$, the policy develops a shortcut behavior, improving rewards by rewriting the prompt description rather than genuinely rectifying the visual content. 
The policy edits the prompt instead of refining the image, which makes the pair appear consistent while the visual error persists. This behavior exposes a critical gap between discriminative verification and constructive rectification. To address this issue, we introduce a second-stage reward that directly encourages effective visual rectification.

\mypara{Stage II: Rectification Alignment Reward.}  In Stage II, the policy generates a rectification action $\hat{a}_i$ for a frozen editor $\Phi$~\cite{qwen_image}. The editor executes $\hat{a}_i$ and produces the refined image $\mathbf{I}^{(1)}$. The policy then re-evaluates the consistency between the original prompt $P$ and $\mathbf{I}^{(1)}$. We define the Rectification Alignment Reward based on the policy's confidence in the consistency of the rectified pair:
\begin{equation}
    R_{\text{rect}} = \mathbb{P}_{\pi_{\theta}}(\hat{v}=\text{True} \mid P, \mathbf{I}^{(1)}).
\end{equation}
By evaluating the rectified pair, $R_{\text{rect}}$ penalizes instruction-based shortcuts and encourages edits that resolve visual inconsistencies, as illustrated in Fig.~\ref{fig:r3_bench}(c). 

Stage I enhances the verifier used to calculate $R_{\text{rect}}$ on $\mathbf{I}^{(1)}$ in Stage II, while Stage II leverages the enhanced verifier to provide execution-grounded supervision, thus promoting effective rectification actions.

\begin{figure}[t]
  \centering
  \includegraphics[width=\linewidth]{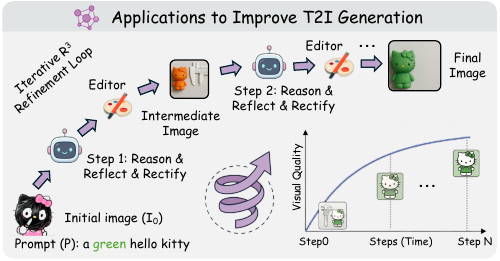} 
\caption{R$^3$-Refiner utilizes the iterative R$^3$ loop to continuously rectify image errors, allowing the final image quality to scale with the number of refinement steps.}
  \label{fig:inference_scaling}
\end{figure}

\mypara{Iterative Refinement with R$^3$-Refiner.}
After training, R$^3$-Refiner employs the R$^3$ loop for iterative refinement in T2I generation. As shown in Fig.~\ref{fig:inference_scaling}, by repeatedly reasoning about image-text alignment and correcting localized errors, the framework progressively enhances image quality with each refinement step. The process continues until the policy confirms consistency or a predefined maximum number of iterations is reached. The effectiveness and generalization of R$^3$-Refiner have been validated across different models, as demonstrated in Tab.~\ref{tab:main_results},~\ref{tab:geneval_results},~\ref{tab:t2i_comp_results}, and~\ref{tab:ablation_bestofn}; iterative improvements are shown in Appendix~\ref{sec:iterative_refinement_analysis}.

\subsection{Scalable Paired Data Construction}
\label{subsec:data_pipeline}
To efficiently train R$^3$-Refiner, we develop a scalable data construction pipeline to obtain paired image-text data consisting of both aligned and misaligned samples, as illustrated in Fig.~\ref{fig:dataset_construct}. 
Each sample is formatted as $\langle P, \mathbf{I}^{(0)}, v_{\text{gt}} \rangle$, where $P$ is the prompt, $\mathbf{I}^{(0)}$ denotes the image, and $v_{\text{gt}} \in \{\text{True}, \text{False}\}$ indicates if $P$ is consistent with $\mathbf{I}^{(0)}$.

\mypara{Multi-Source Synthesis Strategies.}
To address multifaceted alignment challenges, we curate a comprehensive dataset from diverse sources.
To enhance generative quality, our \textit{Generative Ranking} strategy employs a generate-and-rank paradigm based on prompts derived from T2I-R1~\cite{t2i_r1}. Candidate samples are assessed via T2I-CompBench~\cite{t2i_comp}, allowing us to select the top-$k$ ranked samples as positives and the bottom-$k$ as negatives, thus introducing distinct quality differences.

For achieving fine-grained alignment, we adopt the \textit{Counterfactual Rewriting} strategy, which leverages high-quality pairs from BLIP-3O~\cite{blip3onext}. Original pairs are maintained as positive instances, while prompts undergo semantic alterations to intentionally contradict visual content, generating challenging negatives that necessitate precise grounding.
Additionally, to simulate realistic application scenarios, we apply \textit{Visual Inversion} to the PICO-Banana~\cite{pico} dataset. 
Leveraging an MLLM~\cite{qwenvl2_5}, we infer intended prompts based on editing differences, designating the successfully edited images as aligned examples and the pre-edit images as natural negatives.

\mypara{Cascaded Filtering.}
To ensure label fidelity with minimal manual intervention, we introduce a three-stage verification pipeline for cascaded filtering. 
Specifically, the initial stage, \textit{Rationale Verification}, serves as the primary filter employing a Proposer-Verifier mechanism where a specialized MLLM~\cite{Qwen3-VL} generates verdicts and explanations, which a generalist verifier~\cite{omniverifier} subsequently validates to exclude hallucinations lacking visual grounding or logical consistency. 

Then, to further ensure reliability and mitigate model stochasticity, the second stage, \textit{Consensus Voting}, involves repeated querying of a generalist model regarding object presence, quantity, and spatial arrangements. Only instances that consistently achieve high consensus across queries are retained.
Finally, the third stage, \textit{Visual Pruning}, leverages SAM3~\cite{sam3} and CLIP-based scoring to filter out instances with ambiguous object boundaries or insufficient semantic alignment.
Detailed dataset statistics and qualitative comparisons are presented in Appendix~\ref{subsec:filtering_analysis}. The data construction prompts are provided in Appendix~\ref{subsec:data_construction_prompts}.

\begin{table*}[t]
\centering
\caption{\textbf{Quantitative comparison on R$^3$-Bench.} We report Reflective Verdict Score ($\mathcal{S}_{\text{ref}}$) and Rectification Score ($\mathcal{S}_{\text{rect}}$). We adopt Qwen-Image-Edit-2511 as the default editor for standard evaluations. Methods marked with $\dagger$ employ native editing modules. All results are evaluated by Qwen3-VL-235B. \textbf{Bold} indicates the best result within each group.}
\label{tab:main_results}
\small 
\renewcommand{\arraystretch}{1.15} 
\setlength{\tabcolsep}{1.8pt} 

\resizebox{\textwidth}{!}{%
\begin{tabular}{l cc cc cc cc cc cc cc cc cc}
\toprule
\multirow{2.5}{*}{\textbf{Model / Method}} & 
    \multicolumn{2}{c}{\textbf{Color}} & \multicolumn{2}{c}{\textbf{Complex}} & \multicolumn{2}{c}{\textbf{Non-Spa}} & \multicolumn{2}{c}{\textbf{Numeracy}} & \multicolumn{2}{c}{\textbf{Object}} & \multicolumn{2}{c}{\textbf{Shape}} & \multicolumn{2}{c}{\textbf{Spatial}} & \multicolumn{2}{c}{\textbf{Texture}} & \multicolumn{2}{c}{\textbf{Avg}} \\
\cmidrule(lr{0.2em}){2-3} \cmidrule(lr{0.2em}){4-5} \cmidrule(lr{0.2em}){6-7} \cmidrule(lr{0.2em}){8-9} \cmidrule(lr{0.2em}){10-11} \cmidrule(lr{0.2em}){12-13} \cmidrule(lr{0.2em}){14-15} \cmidrule(lr{0.2em}){16-17} \cmidrule(lr{0.2em}){18-19}
& $\mathcal{S}_{\text{ref}}$ & $\mathcal{S}_{\text{rect}}$ & $\mathcal{S}_{\text{ref}}$ & $\mathcal{S}_{\text{rect}}$ & $\mathcal{S}_{\text{ref}}$ & $\mathcal{S}_{\text{rect}}$ & $\mathcal{S}_{\text{ref}}$ & $\mathcal{S}_{\text{rect}}$ & $\mathcal{S}_{\text{ref}}$ & $\mathcal{S}_{\text{rect}}$ & $\mathcal{S}_{\text{ref}}$ & $\mathcal{S}_{\text{rect}}$ & $\mathcal{S}_{\text{ref}}$ & $\mathcal{S}_{\text{rect}}$ & $\mathcal{S}_{\text{ref}}$ & $\mathcal{S}_{\text{rect}}$ & $\mathcal{S}_{\text{ref}}$ & $\mathcal{S}_{\text{rect}}$ \\
\midrule
GPT-4o~\cite{gpt4o} & 0.77 & 0.64 & 0.72 & 0.30 & 0.80 & 0.48 & 0.78 & 0.41 & 0.87 & 0.87 & 0.74 & 0.54 & 0.72 & 0.41 & 0.71 & 0.53 & 0.76 & 0.53 \\
Banana & 0.84 & 0.67 & 0.78 & 0.31 & 0.78 & 0.07 & 0.90 & 0.49 & 0.92 & 0.78 & 0.83 & 0.39 & 0.88 & 0.48 & 0.80 & 0.55 & 0.84 & 0.50 \\
GPT-Image-1 & 0.82 & 0.68 & 0.80 & 0.39 & 0.78 & 0.33 & 0.74 & 0.52 & \textbf{0.96} & \textbf{1.00} & 0.74 & 0.46 & 0.84 & \textbf{0.63} & 0.76 & 0.58 & 0.79 & 0.57 \\
Gemini-3-Pro & 0.85 & \textbf{0.73} & 0.81 & 0.56 & 0.69 & 0.43 & \textbf{0.95} & 0.39 & \textbf{0.96} & 0.94 & 0.83 & 0.50 & \textbf{0.93} & 0.56 & 0.86 & 0.60 & \textbf{0.87} & 0.60 \\
GPT-5.2 & 0.82 & 0.71 & 0.63 & \textbf{0.57} & 0.61 & 0.86 & 0.91 & \textbf{0.61} & 0.92 & 0.96 & 0.71 & \textbf{0.56} & 0.89 & 0.55 & 0.84 & \textbf{0.63} & 0.80 & \textbf{0.65} \\
\midrule
Bagel$^{\dagger}$~\cite{bagel} & 0.57 & 0.33 & 0.57 & 0.07 & 0.90 & 0.00 & 0.46 & 0.05 & 0.63 & 0.46 & 0.45 & 0.19 & 0.45 & 0.12 & 0.66 & 0.34 & 0.56 & 0.21 \\
OmniGen2$^{\dagger}$~\cite{omnigen2} & 0.58 & -0.09 & 0.43 & -0.44 & 0.41 & 0.14 & 0.55 & -0.39 & 0.49 & 0.08 & 0.53 & -0.27 & 0.53 & -0.18 & 0.50 & -0.09 & 0.51 & -0.19 \\
ReasonEdit$^{\dagger}$~\cite{reasonedit} & 0.49 & 0.19 & 0.41 & -0.02 & 0.76 & 0.00 & 0.60 & 0.10 & 0.56 & 0.14 & 0.50 & 0.09 & 0.55 & -0.03 & 0.54 & 0.11 & 0.54 & 0.08 \\
SLD~\cite{sld} & 0.44 & 0.17 & 0.21 & 0.16 & 0.14 & -0.24 & 0.51 & 0.34 & 0.28 & 0.16 & 0.24 & 0.18 & 0.57 & 0.52 & 0.39 & 0.41 & 0.37 & 0.28 \\
UniCot$^{\dagger}$~\cite{unicot} & 0.67 & 0.38 & 0.66 & 0.27 & 0.88 & 0.29 & 0.74 & 0.29 & 0.68 & 0.51 & 0.62 & 0.23 & 0.63 & 0.21 & 0.69 & 0.38 & 0.68 & 0.32 \\
ReflectionFlow$^{\dagger}$~\cite{reflectionflow} & 0.67 & 0.18 & 0.77 & -0.03 & 0.80 & -0.07 & 0.77 & 0.28 & 0.85 & 0.61 & 0.80 & 0.06 & 0.79 & 0.32 & 0.80 & 0.45 & 0.78 & 0.26 \\
Reflect-DiT$^{\dagger}$~\cite{reflectdit} & 0.73 & 0.25 & 0.53 & 0.25 & 0.92 & 0.57 & 0.51 & 0.37 & 0.82 & 0.76 & 0.64 & 0.08 & 0.75 & 0.39 & 0.64 & 0.53 & 0.68 & 0.37 \\
ThinkGen$^{\dagger}$~\cite{thinkgen} & 0.78 & 0.47 & \textbf{0.84} & 0.07 & 0.90 & 0.14 & 0.92 & 0.32 & 0.93 & 0.65 & \textbf{0.86} & 0.25 & 0.89 & 0.38 & \textbf{0.89} & 0.43 & \textbf{0.87} & 0.37 \\
OmniVerifier~\cite{omniverifier} & 0.80 &  0.28 & 0.71 & -0.18 & 0.92 & 0.62 & 0.84 & 0.24 & 0.87 & 0.14 & 0.80 & 0.08 & 0.76 & 0.17 & 0.76 & 0.31 & 0.80 & 0.17 \\

\midrule[\heavyrulewidth]
Qwen2.5-VL-7B~\cite{qwenvl2_5} & 0.70 & 0.56 & 0.68 & 0.20 & 0.76 & 0.86 & 0.74 & 0.23 & 0.80 & 0.71 & 0.71 & 0.31 & 0.67 & 0.24 & 0.76 & 0.45 & 0.72 & 0.38 \\
\rowcolor{gray!15} \hspace{0.5em}+ R$^3$-Refiner (Ours) & \textbf{0.86} & 0.67 & 0.81 & 0.24 & \textbf{0.94} & 0.71 & 0.85 & 0.37 & 0.86 & 0.68 & 0.82 & 0.43 & 0.88 & 0.38 & 0.80 & 0.51 & 0.84 & 0.47 \\
\addlinespace[0.2em]

Qwen3-VL-8B~\cite{Qwen3-VL} & 0.77 & 0.64 & 0.71 & 0.46 & 0.63 & 0.57 & 0.83 & 0.45 & 0.88 & 0.95 & \textbf{0.86} & 0.43 & 0.83 & 0.40 & 0.77 & 0.55 & 0.80 & 0.54 \\
\rowcolor{gray!15} \hspace{0.5em}+ R$^3$-Refiner (Ours) & 0.81 & 0.72 & 0.82 & 0.33 & 0.92 & \textbf{1.00} & 0.86 & 0.52 & 0.92 & 0.94 & \textbf{0.86} & \textbf{0.56} & 0.92 & 0.62 & 0.83 & 0.56 & \textbf{0.87} & 0.62  \\

\bottomrule
\end{tabular}%
}
\vspace{-0.5em} 
\end{table*}

\begin{table}[t]
\centering
\caption{\textbf{Quantitative results on GenEval++.} Each model uses the same model for initial generation and subsequent editing, except that Qwen-Image uses Qwen-Image for generation and Qwen-Image-Edit for editing. R$^3$-Refiner performs verification and provides rectification instructions. All results are evaluated by Qwen3-VL-235B.}
\label{tab:geneval_results}
\small
\renewcommand{\arraystretch}{1.1}
\setlength{\tabcolsep}{2.8pt}
\resizebox{\columnwidth}{!}{%
\begin{tabular}{l cccccccc}
\toprule
\textbf{Model} & \textbf{Color} & \textbf{Count} & \textbf{Col/Cnt} & \textbf{Col/Pos} & \textbf{Pos/Cnt} & \textbf{Pos/Size} & \textbf{Multi} & \textbf{Avg} \\
\midrule
SD-3-Med~\cite{sd3} & 0.550 & 0.500 & 0.125 & 0.350 & 0.175 & 0.150 & 0.225 & 0.296 \\
FLUX.1~\cite{flux}& 0.350 & 0.625 & 0.150 & 0.275 & 0.200 & 0.375 & 0.225 & 0.314 \\
Janus-Pro~\cite{janus_pro} & 0.450 & 0.300 & 0.125 & 0.300 & 0.075 & 0.350 & 0.125 & 0.246 \\
\midrule
Bagel~\cite{bagel} & 0.575 & 0.500 & 0.350 & 0.300 & 0.175 & 0.625 & 0.425 & 0.421 \\
\rowcolor{gray!15} \hspace{0.5em}+ R$^3$-Refiner & 0.650 & 0.650 & 0.500 & 0.400 & 0.300 & 0.675 & 0.550 & 0.532 \\
\addlinespace[2pt]

OmniGen2~\cite{omnigen2} & 0.625 & 0.250 & 0.150 & 0.300 & 0.100 & 0.475 & 0.300 & 0.314 \\
\rowcolor{gray!15} \hspace{0.5em}+ R$^3$-Refiner & 0.650 & 0.300 & 0.325 & 0.300 & 0.100 & 0.525 & 0.350 & 0.364 \\
\addlinespace[2pt]

Qwen-Image~\cite{qwen_image} & 0.800 & 0.700 & 0.600 & 0.700 & 0.500 & 0.725 & 0.550 & 0.654 \\
\rowcolor{gray!15} \hspace{0.5em}+ R$^3$-Refiner & 0.925 & 0.775 & 0.775 & 0.725 & 0.550 & 0.700 & 0.550 & 0.714 \\

\midrule
GPT Image & 0.925 & \textbf{0.900} & 0.825 & 0.625 & \textbf{0.600} & \textbf{0.875} & 0.800 & 0.793 \\
\rowcolor{gray!15} \hspace{0.5em}+ R$^3$-Refiner & \textbf{0.950} & \textbf{0.900} & \textbf{0.925} & 0.675 & 0.575 & \textbf{0.875} & 0.900 & \textbf{0.829} \\
\addlinespace[2pt]

Banana & 0.875 & 0.775 & 0.625 & 0.700 & 0.500 & 0.775 & 0.900 & 0.736 \\
\rowcolor{gray!15} \hspace{0.5em}+ R$^3$-Refiner & 0.925 & 0.850 & 0.725 & \textbf{0.775} & 0.575 & 0.775 & \textbf{0.975} & 0.800 \\
\bottomrule
\end{tabular}%
}
\end{table}

\section{Experiments}
\label{sec:experiments}
In this section, we present the main results of R$^3$-Refiner on R$^3$-Bench and compare it against representative state-of-the-art verification and refinement methods. Subsequently, we demonstrate the plug-and-play capabilities of R$^3$-Refiner on general T2I benchmarks and analyze design choices through ablation studies.

\subsection{Benchmark Results}
\label{subsec:benchmark_results}

\mypara{Results on R$^3$-Bench.}
Tab.~\ref{tab:main_results} reports the quantitative results on R$^3$-Bench. We evaluate state-of-the-art (SOTA) models across the following categories: UMMs (Bagel, OmniGen2), MLLMs (Qwen2.5-VL, Qwen3-VL), existing RVG methods (SLD, ReasonEdit, UniCot, ReflectionFlow, Reflect-DiT, ThinkGen, OmniVerifier), and closed-source models (Gemini 3, GPT-4o, GPT-5.2, Banana, GPT-Image-1). R$^3$-Refiner achieves SOTA performance among open-source methods. Specifically, our method (built on Qwen3-VL-8B) attains an $\mathcal{S}_{\text{ref}}$ of 0.87, matching the performance of powerful closed-source models such as Gemini 3. In terms of rectification efficacy ($\mathcal{S}_{\text{rect}}$), while GPT-5.2 leads with 0.65, R$^3$-Refiner yields a competitive 0.62, demonstrating that our RL-based optimization can effectively distill reasoning capabilities into effective rectification actions. Appendix~\ref{subsec:benchmark_reliability} further verifies the reliability of these comparisons with bootstrap and rank-stability analyses, and Appendix~\ref{subsec:cross_editor_generalization} reports additional training-editor transfer results on R$^3$-Bench under multiple inference-time editors.

\begin{table}[t]
\centering
\caption{\textbf{Quantitative results on T2I-CompBench.} We follow the same generation and editing setup as in Tab.~\ref{tab:geneval_results}.}
\label{tab:t2i_comp_results}
\small
\renewcommand{\arraystretch}{1.1}
\setlength{\tabcolsep}{3pt}

\resizebox{\columnwidth}{!}{%
\begin{tabular}{l cccccc}
\toprule
\textbf{Method} & \textbf{Color} & \textbf{Shape} & \textbf{Texture} & \textbf{Spatial} & \textbf{Complex} & \textbf{Avg} \\
\midrule
PixArt-$\alpha$~\cite{pixart}  & 0.669 & 0.493 & 0.648 & 0.206 & 0.343 & 0.472 \\
SD-v1.5~\cite{ldm}  & 0.376 & 0.371 & 0.419 & 0.117 & 0.305 & 0.318 \\
SD-XL-base-1.0~\cite{sdxl}  & 0.588 & 0.469 & 0.530 & 0.213 & 0.324 & 0.425 \\
FLUX.1~\cite{flux}  & 0.741 & 0.572 & 0.692 & 0.286 & 0.370 & 0.532 \\
Janus-Pro~\cite{janus_pro} & 0.636 & 0.353 & 0.494 & 0.206 & 0.356 & 0.409 \\
T2I-R1~\cite{t2i_r1} & 0.813 & 0.585 & 0.724 & 0.338 & \textbf{0.399} & 0.572 \\
\midrule

OmniGen2~\cite{omnigen2} & 0.776 & 0.516 & 0.709 & 0.393 & 0.371 & 0.553 \\
\rowcolor{gray!15} \hspace{0.5em}+ R$^3$-Refiner & 0.801 & 0.521 & 0.721 & \textbf{0.401} & 0.378 & 0.564 \\
\addlinespace[3pt]

Bagel~\cite{bagel} & 0.796 & 0.571 & 0.686 & 0.327 & 0.386 & 0.553 \\
\rowcolor{gray!15} \hspace{0.5em}+ R$^3$-Refiner & \textbf{0.842} & \textbf{0.603} & \textbf{0.739} & 0.357 & 0.398 & \textbf{0.588} \\
\bottomrule
\end{tabular}%
}
\vspace{-1em}
\end{table}

\begin{figure*}[t]
\centering
\includegraphics[width=\textwidth]{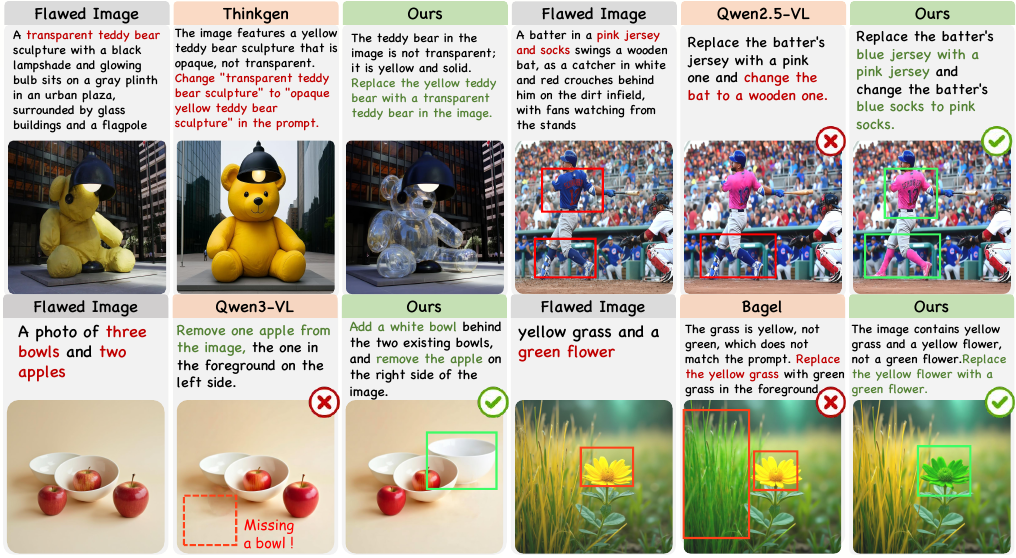}
\caption{Qualitative comparison of R$^3$-Refiner with existing MLLMs, UMMs, and RVG methods.}
\label{fig:vis}
\end{figure*}

\mypara{Results on T2I Benchmarks.}
To assess the generalization capability of R$^3$-Refiner, we evaluate its performance as a plug-and-play module on standard T2I benchmarks, including GenEval++~\cite{geneval++} and T2I-CompBench~\cite{t2i_comp}. We employ the iterative refinement loop described in Sec.~\ref{subsec:r3_refiner} with a maximum of two iterations. On GenEval++, R$^3$-Refiner consistently improves diverse base generators, including Bagel and Qwen-Image, and also enhances strong closed-source models such as Banana and GPT Image. For GenEval++, Appendix~\ref{subsec:cross_editor_generalization} reports additional training-editor transfer results under multiple inference-time editors. We observe similar trends on T2I-CompBench, where R$^3$-Refiner improves the average scores of both OmniGen2 and Bagel. These results confirm that our policy improves visual generation through iterative refinement.

\begin{table}[t]
\centering
\caption{Ablation Study on Rectification Alignment Reward.}
\label{tab:ablation_reward}
\renewcommand{\arraystretch}{1.2} 
\setlength{\tabcolsep}{6pt} 
\resizebox{\columnwidth}{!}{
\begin{tabular}{l cc}
\toprule
Reward Mechanism & Reasoning ($S_{\text{ref}}$) & Rectification ($S_{\text{rect}}$) \\
\midrule
Qwen2.5-VL-7B~\cite{qwenvl2_5} & 0.72 & 0.38 \\
Reasoning Reward Only ($R_{\text{reason}}$) & 0.81\inc{0.09} & 0.29\dec{0.09} \\
Hybrid ($R_{\text{reason}}$ + Decomposed QA) & 0.85\inc{0.13} & 0.41\inc{0.03} \\
Hybrid ($R_{\text{reason}}$ + SAM3 + CLIP) & 0.84\inc{0.12} & 0.35\dec{-0.03} \\
\midrule
\rowcolor{gray!15} HRM (Ours) & 0.84\inc{0.12} & \textbf{0.47}\inc{0.09} \\
\bottomrule
\end{tabular}%
}
\vspace{-1em}
\end{table}

\subsection{Ablation Study}
\label{subsec:ablation_study}

\mypara{Rectification Alignment Reward Design.}
The proposed R$^3$-Refiner is a two-stage reinforcement learning framework that utilizes GT labels for the first-stage reward. We investigate multiple alternatives for the second-stage reward design. Following~\cite{t2i_r1}, we employ a CLIP-detector pipeline to generate fine-grained reward signals, substituting the detector with SAM3. Additionally, we analyze question decomposition by partitioning prompt elements into sub-questions and calculating individual rewards via VQA. As shown in Tab.~\ref{tab:ablation_reward}, applying only the first-stage reward causes the model to exhibit illusory visual rectification (Sec.~\ref{subsec:r3_refiner}) and leads to rectification score degeneration. In contrast, the simplest image-text matching reward mechanism achieves optimal performance, demonstrating the effectiveness of the proposed self-reward paradigm.

\begin{table}[t]
\centering
\caption{Comparison of R$^3$-Refiner and Best-of-$N$ on GenEval++.}
\label{tab:ablation_bestofn}
\small
\renewcommand{\arraystretch}{1.1}
\setlength{\tabcolsep}{2.8pt}
\resizebox{\columnwidth}{!}{%
\begin{tabular}{l c cccccccc}
\toprule
\textbf{Model} & $N$ & \textbf{Color} & \textbf{Count} & \textbf{Col/Cnt} & \textbf{Col/Pos} & \textbf{Pos/Cnt} & \textbf{Pos/Size} & \textbf{Multi} & \textbf{Avg} \\
\midrule
\addlinespace[2pt]
Qwen-Image~\cite{qwen_image} & 0 & 0.800 & 0.700 & 0.600 & 0.700 & 0.500 & 0.725 & \textbf{0.550} & 0.654 \\
\hspace{0.5em}+ Best-of-N~\cite{qwenvl2_5} & 3 & 0.875 & 0.700 & 0.725 & 0.700 & 0.550 & 0.725 & 0.500 & 0.682 \\
\hspace{0.5em}+ Best-of-N & 4 & 0.825 & 0.725 & 0.625 & 0.725 & \textbf{0.575} & 0.725 & 0.500 & 0.671 \\
\hspace{0.5em}+ Best-of-N & 5 & 0.850 & 0.725 & 0.625 & \textbf{0.750} & 0.550 & \textbf{0.750} & \textbf{0.550} & 0.686 \\
\hspace{0.5em}+ Best-of-N & 6 & 0.825 & 0.725 & 0.600 & \textbf{0.750} & \textbf{0.575} & 0.725 & \textbf{0.550} & 0.679 \\
\midrule
\hspace{0.5em}+ R$^3$-Refiner & 1 & \textbf{0.925} & \textbf{0.800} & 0.650 & 0.725 & \textbf{0.575}  & 0.700 & \textbf{0.550} & 0.703 \\
\rowcolor{gray!15}
\hspace{0.5em}+ R$^3$-Refiner & 2 & \textbf{0.925} & 0.775 & \textbf{0.775} & 0.725 & 0.550 & 0.700 & \textbf{0.550} & \textbf{0.714} \\
\hspace{0.5em}+ R$^3$-Refiner & 3 & \textbf{0.925} & \textbf{0.800} &0.650  &0.700  &\textbf{0.575}  &0.700  & \textbf{0.550} &0.700  \\
\bottomrule
\end{tabular}%
}
\end{table}

\mypara{Comparison to Best-of-N.} To further enhance image generation quality, another strategy commonly employed during the inference phase is ``Best-of-$N$". This strategy generates $N$ candidate images using different random seeds and subsequently utilizes an external evaluator to select the highest quality sample. This method improves quality at the cost of increased parallel computational overhead, contrasting with the serial optimization paradigm of R$^3$-Refiner. We compare the performance of R$^3$-Refiner against this strategy on the GenEval++ dataset, as presented in Tab.~\ref{tab:ablation_bestofn}. Experimental results indicate that R$^3$-Refiner outperforms the peak performance of the Best-of-$N$ approach with a single RVG iteration. Furthermore, we observed a performance saturation phenomenon in both methods: as the value of $N$ increases, the performance of the Best-of-$N$ method does not improve significantly. Similarly, the performance of R$^3$-Refiner tends to saturate after two refinement iterations.

\mypara{Human Evaluation.}
To validate the automatic rectification metric, we conduct a human study on 24 category-balanced R$^3$-Bench instances with 23 annotators. Annotators answer factual yes/no questions derived from the original prompts, and we compare the resulting human QA accuracy with $\mathcal{S}_{\text{rect}}$ over four representative models. R$^3$-Refiner-BG denotes the variant trained with Bagel~\cite{bagel}. As shown in Tab.~\ref{tab:human_eval}, human judgments preserve the same coarse ordering as the automatic metric, with GPT-5.2 and R$^3$-Refiner-BG tied at the top. The two rankings are strongly aligned (SROCC=0.800, KROCC=0.667), and annotators show consistent agreement (Fleiss' $\kappa=0.776$). Additional evaluator-swap results are provided in Appendix~\ref{subsec:evaluator_robustness}.

\mypara{Iterative Inference vs. Learned Policy.}
To separate the effect of iterative editing from policy learning, we compare R$^3$-Refiner with two non-learned iterative alternatives under the same Qwen-Image-Edit editor~\cite{qwen_image}. Prompt resubmission re-feeds the original prompt and edited image to the editor without verification. R$^3$-Refiner-BG denotes the variant trained with Bagel~\cite{bagel}, and the pretrained verifier uses Qwen3-VL-8B~\cite{Qwen3-VL}. As shown in Tab.~\ref{tab:iterative_policy}, prompt resubmission improves slightly at first but drops back by round two, suggesting that repeated editing without verification can corrupt already aligned content. The pretrained verifier degrades with iteration due to excessive false positives in verification. In contrast, only R$^3$-Refiner variants improve consistently across rounds, supporting that gains stem from the learned policy.

\begin{table}[t]
\centering
\caption{\textbf{Human validation of $\mathcal{S}_{\text{rect}}$.} Human QA accuracy is compared with the automatic rectification score across representative models.}
\label{tab:human_eval}
\small
\renewcommand{\arraystretch}{1.1}
\setlength{\tabcolsep}{3.2pt}
\resizebox{\columnwidth}{!}{%
\begin{tabular}{l cccc}
\toprule
\textbf{Metric} & \textbf{GPT-5.2} & \textbf{R$^3$-Refiner-BG} & \textbf{Gemini-3-Pro} & \textbf{Qwen3-VL-8B} \\
\midrule
$\mathcal{S}_{\text{rect}}$ & 0.650 & 0.657 & 0.599 & 0.543 \\
Human QA Acc. & 0.912 & 0.912 & 0.829 & 0.719 \\
\bottomrule
\end{tabular}%
}
\vspace{-0.75em}
\end{table}

\section{Conclusion}
\label{sec:conclusion}

We propose R$^3$-Refiner to advance Reflective Visual Generation by addressing the misalignment where MLLMs accurately diagnose errors but fail to execute valid corrections. By incorporating a Hierarchical Reward Mechanism, our approach aligns the Iterative R$^3$ loop to facilitate precise and progressive visual refinement. Experiments on R$^3$-Bench, GenEval++, and T2I-CompBench demonstrate that our policy outperforms rigid verifiers and functions as a robust plug-and-play module for diverse generative executors. These findings highlight the value of Inference-Time Scaling for reliable visual synthesis.

\begin{table}[t]
\centering
\caption{\textbf{Iterative inference vs. learned policy on GenEval++.} All methods use Qwen-Image-Edit as the editor and report the average score.}
\label{tab:iterative_policy}
\small
\renewcommand{\arraystretch}{1.1}
\setlength{\tabcolsep}{3.4pt}
\resizebox{\columnwidth}{!}{%
\begin{tabular}{l cccc}
\toprule
\textbf{Method} & \textbf{Policy Source} & \textbf{$N=0$} & \textbf{$N=1$} & \textbf{$N=2$} \\
\midrule
Qwen-Image & -- & 0.654 & -- & -- \\
\hspace{0.5em}+ Prompt resubmission & None & 0.654 & 0.668 & 0.654 \\
\hspace{0.5em}+ Qwen3-VL-8B verifier & Pretrained & 0.654 & 0.639 & 0.571 \\
\hspace{0.5em}+ R$^3$-Refiner-QE & RL, Qwen-Edit & 0.654 & 0.704 & \textbf{0.714} \\
\rowcolor{gray!15}
\hspace{0.5em}+ R$^3$-Refiner-BG & RL, Bagel & 0.654 & 0.686 & 0.711 \\
\bottomrule
\end{tabular}%
}
\vspace{-0.75em}
\end{table}

\section*{Acknowledgement}
This work was supported by the Guangdong Basic and Applied Basic Research Foundation (2025A1515011546) and by the Shenzhen Science and Technology Program (JCYJ20240813105901003, ZDCY20250901113000001).

\section*{Impact Statement}
This paper introduces R$^3$-Bench and R$^3$-Refiner to improve the reliability of visual generative models by helping them diagnose and correct visual errors. Potential positive impacts include reducing misaligned generated content in creative, educational, and assistive applications. Potential risks include enabling more capable image-generation and editing systems that could be misused to create misleading synthetic content. We encourage deployment with provenance tracking, watermarking, access controls, and safeguards aligned with applicable policies.

\bibliography{main}
\bibliographystyle{icml2026}
\appendix
\ifdefined\maketitlesupplementary
    \maketitlesupplementary
\else
    \twocolumn[{
        \centering
        \Large \textbf{Supplementary Material} \\
        \vspace{1.0em}
    }]
\fi

{
    \hypersetup{linkcolor=black}
    \setlength{\parskip}{2pt}
    \AppTOCTitle
    \apptocA{sec:related_work}{A \quad Related Work}

    \apptocA{sec:qual_results}{B \quad Additional Qualitative Results}
    \apptocB{subsec:filtering_analysis}{B.1 \quad Data Filtering Analysis}
    \apptocB{subsec:r3_vis}{B.2 \quad Extended Visualization of R$^3$-Bench}
    \apptocB{subsec:failure_case}{B.3 \quad Failure Case Analysis}

    \apptocA{sec:additional_quant_analysis}{C \quad Additional Quantitative Analysis}
    \apptocB{subsec:evaluator_robustness}{C.1 \quad Evaluator Robustness}
    \apptocB{subsec:cross_editor_generalization}{C.2 \quad Training-Editor Transfer}
    \apptocB{subsec:benchmark_reliability}{C.3 \quad Benchmark Reliability}
    \apptocB{sec:iterative_refinement_analysis}{C.4 \quad Iterative Refinement Analysis}

    \apptocA{sec:train_details}{D \quad Training Implementation Details}

    \apptocA{sec:metrics}{E \quad Evaluation Metrics Details}
    \apptocB{subsec:reflect_score}{E.1 \quad Reflective Verdict Score ($\mathcal{S}_{\text{ref}}$)}
    \apptocB{subsec:rectify_score}{E.2 \quad Rectification Score ($\mathcal{S}_{\text{rect}}$)}

    \apptocA{sec:prompts}{F \quad Prompt Details}
    \apptocB{subsec:training_prompt}{F.1 \quad Training Prompt for R$^3$-Refiner}
    \apptocB{subsec:eval_prompts}{F.2 \quad Evaluation Prompts}
    \apptocB{subsec:data_construction_prompts}{F.3 \quad Data Construction Prompts}
    \apptocA{sec:test_set_details}{G \quad Details of Test Set Curation}
}

\section{Related Work}
\label{sec:related_work}
\subsection{Text-to-Image (T2I) Generation}
T2I generation has advanced significantly. Leading diffusion models, such as Stable Diffusion~\cite{sd3, sdxl} and FLUX~\cite{flux, flux1kontext}, demonstrate impressive generative capabilities through large-scale training. Recent research shifts toward UMMs~\cite{dim, mmada_parallel, bridge, harmon, openuni, x_omni, illume+, infinitystar, mogao, lightbagel, qwen_image, mentor, tbac, blip3onext}. These UMMs build upon the reasoning capabilities of MLLMs~\cite{blip2, kimi, deepseekvl, llava, keyevl1_5, llavaoneversion, qwenvl2_5, Qwen3-VL, internvl3_5, llama_adapter, minigpt4} and integrate multimodal understanding with generation into a unified architecture for controllable synthesis. Representative works include Emu~\cite{emu3_5}, Show-o2~\cite{showo2}, Janus-Pro~\cite{janus_pro}, OmniGen2~\cite{omnigen2}, and Lumina-DiMOO~\cite{lumina_dimoo}. For instance, Bagel~\cite{bagel} and MMaDA~\cite{mmada} utilize large-scale interleaved multimodal data and exhibit emergent capabilities in complex generation and reasoning. Simultaneously, Z-Image~\cite{zimage} focuses on efficient native generation architectures. Despite these advancements, these models still struggle with compositional prompts as they operate in an open-loop and single-pass paradigm. This approach lacks mechanisms for error rectification and necessitates a transition to a multi-round RVG paradigm.

\subsection{Reasoning and Reflection in Visual Generation}
Inspired by the success of self-reflection mechanisms in LLMs~\cite{react, reflexion, Cot_valve, mitigating, seal, reasoning, fractional, thinking, dast, think, thinkless} and MLLMs~\cite{sherlock, critic, llava_critic_r1, self_correct, self_refine, prometheus}, recent visual generation studies~\cite{mure, ThinkMorph} explore reasoning generation and closed-loop paradigms. Several approaches~\cite{thinkgen, dim, reasonedit} employ chain-of-thought reasoning to optimize input prompts and guide the image generation and editing process. ThinkMorph~\cite{ThinkMorph} investigates interleaved multimodal reasoning to align semantic understanding with visual synthesis. SLD~\cite{sld} and OmniVerifier~\cite{omniverifier} serve as plug-and-play verifiers that detect and correct errors in image generation. Other strategies~\cite{unicot, IRG, reflectionflow, reflectdit} utilize model-generated critiques to guide iterative refinements for enhanced semantic alignment and visual fidelity. However, we identify a critical capability misalignment where models fail to translate diagnostic reasoning into effective correction. Consequently, we propose a general reinforcement learning framework that aligns discriminative capabilities with actionable rectification by optimizing the entire Reason-Reflect-Rectify loop.

\subsection{Benchmarks for Visual Generation and Verification}
Existing benchmarks primarily evaluate isolated capabilities within the domains of visual generation~\cite{geneval++, tiif, rise, t2i_reasonbench} and verification~\cite{omniverifier}. For instance, T2I-CompBench~\cite{t2i_comp, t2i_comp_plus}, GenEval~\cite{geneval}, and DPG-Bench~\cite{dpgbench} target attribute alignment and compositional generation tasks, including spatial relationship modeling. Similarly, WISE~\cite{wise} and KRIS~\cite{kris} assess the integration of world knowledge and commonsense reasoning into visual generation and editing. Furthermore, OneIG~\cite{oneig}, MME-Unify~\cite{mme_unify}, and RealUnify~\cite{realunify} introduce unified architectures covering understanding, generation, and multimodal tasks. However, these benchmarks predominantly focus on open-loop evaluation and fail to quantify the iterative reasoning integral to Reflective Visual Generation. To bridge this gap, we introduce R$^3$-Bench, which formalizes the Reason-Reflect-Rectify loop to assess the alignment between diagnostic reasoning and actionable rectification.

\section{Additional Qualitative Results}
\label{sec:qual_results}

\subsection{Data Filtering Analysis}
\label{subsec:filtering_analysis}

\begin{figure}[t]
    \centering
    \includegraphics[width=.9\columnwidth]{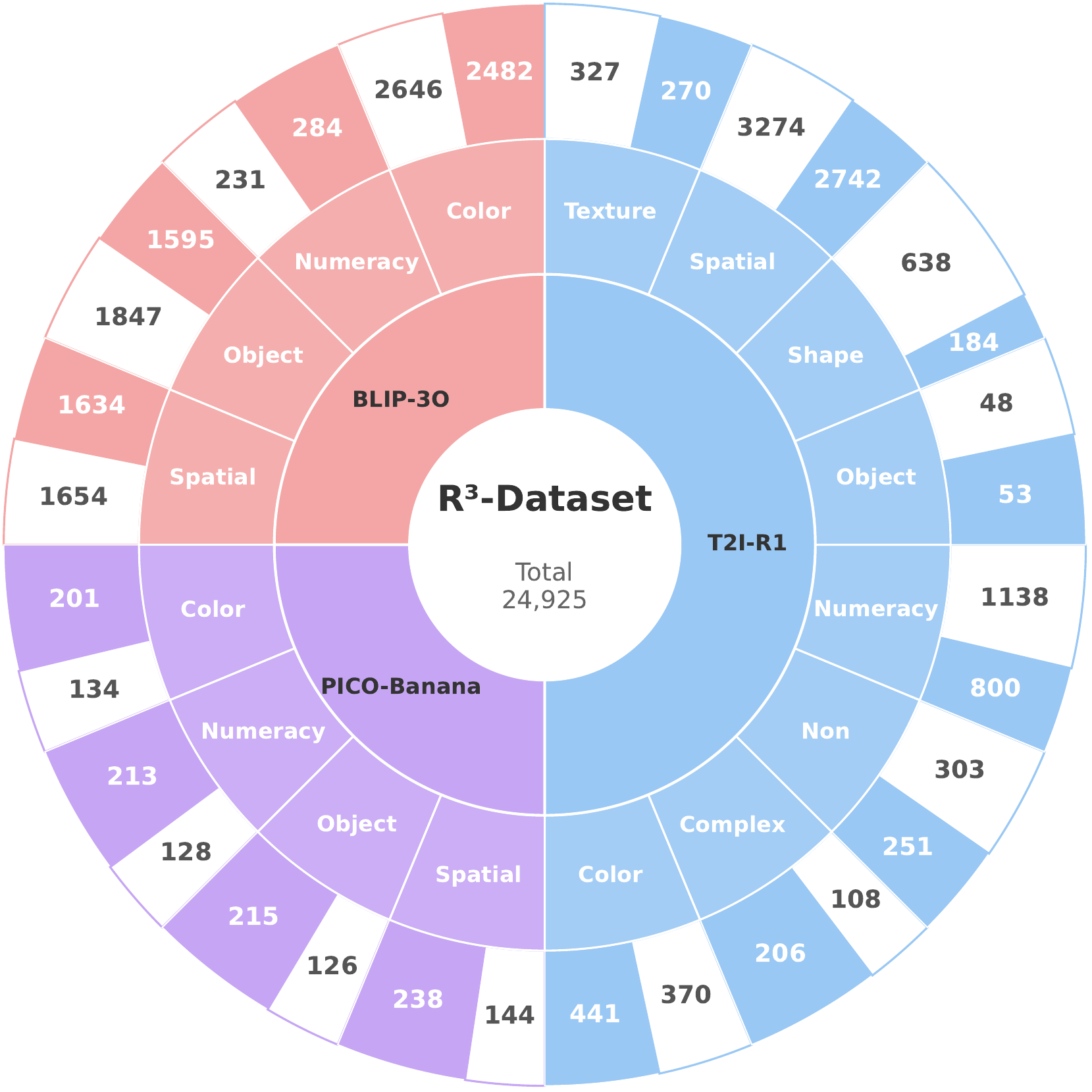}
    \caption{\textbf{Hierarchical Distribution of the R$^3$-Dataset.} 
    The inner ring displays data sources, including T2I-R1, BLIP-3O, and PICO-Banana. Moving outward, the middle ring illustrates fine-grained error categories, such as Spatial, Color, and Numeracy. These categories are distributed evenly to maintain balanced visual diversity. Finally, the outer ring details the composition of the final preference pairs by indicating specific counts of Aligned (solid) and Misaligned (hollow) samples. The consistent presence of high-quality hard negatives alongside positives across all categories demonstrates the efficacy of the proposed data construction strategies.}
    \label{fig:dataset_sunburst}
\end{figure}

\begin{figure}[t]
    \centering
    \includegraphics[width=\columnwidth]{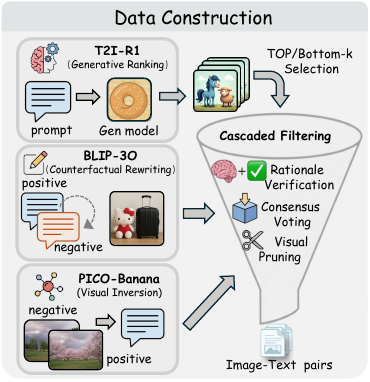}
    \caption{\textbf{Overview of the Scalable Paired Data Construction Pipeline.} 
    The proposed pipeline is structured into two primary phases. The Multi-Source Synthesis Strategies phase initially leverages Generative Ranking, Counterfactual Rewriting, and Visual Inversion to synthesize diverse candidates with hard negatives. Subsequently, the Cascaded Filtering phase implements a three-stage verification mechanism comprising Rationale Verification, Consensus Voting, and Visual Pruning. This rigorous validation ensures high label fidelity and eliminates visual hallucinations. Ultimately, the pipeline yields high-quality aligned and misaligned sample pairs.}
    \label{fig:dataset_construct}
\end{figure}

To evaluate the efficacy of the proposed Automated Cascaded Filtering pipeline (Fig.~\ref{fig:dataset_construct}), we present a detailed statistical breakdown of the dataset composition alongside qualitative comparisons between rejected noise and the final high-quality data.
\paragraph{Dataset Composition.}
From an initial pool of approximately 40,000 synthesized samples, the three-stage filtering pipeline yielded a final R$^3$-Dataset of 24,925 high-fidelity instances. Fig.~\ref{fig:dataset_sunburst} depicts the hierarchical distribution of the curated dataset. This visualization details contributions from three distinct sources (inner ring), the diversity of fine-grained categories (middle ring), and the composition of preference pairs (outer ring). Specifically, the outer ring presents the distribution of aligned (positive) versus misaligned (negative) samples. This balanced structure facilitates model learning in distinguishing correct visual depictions from subtle hallucinations.

\begin{figure*}[t]
    \centering
    \includegraphics[width=\textwidth]{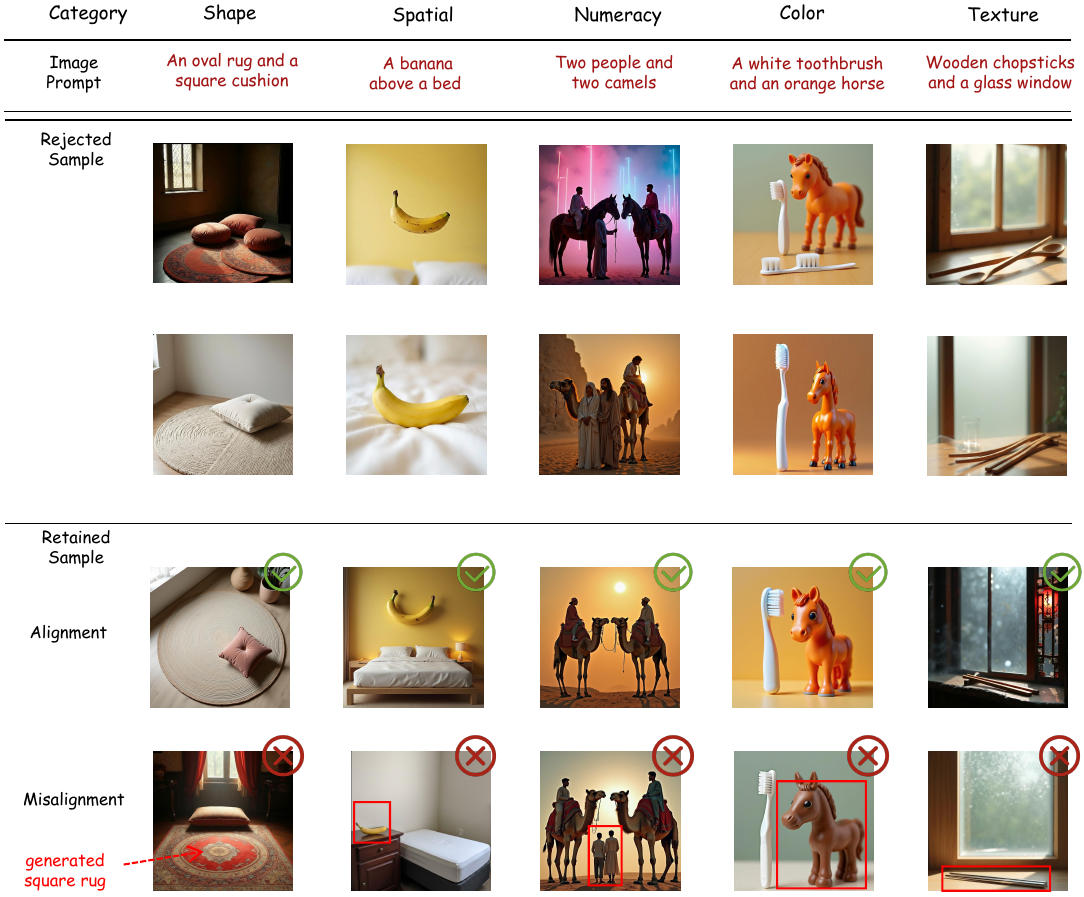}
    \caption{\textbf{Qualitative comparison of filtered noise and final data.} \textbf{Top:} Low-quality samples rejected by our pipeline due to hallucinations, ambiguity, or visual artifacts. \textbf{Bottom:} High-quality retained preference pairs. Each pair consists of an \textbf{Aligned} image (matching the prompt) and a \textbf{Misaligned} image (containing specific errors), providing the contrastive signal needed for training.}
    \label{fig:filtering_qualitative}
\end{figure*}

\mypara{Qualitative Quality Control.}
Fig.~\ref{fig:filtering_qualitative} illustrates the efficacy of our quality control process. The top row displays rejected instances discarded due to critical deficiencies such as logical hallucinations (text contradicting image content), visual ambiguity, or segmentation artifacts. Conversely, the bottom row presents retained high-quality preference pairs that satisfy all verification criteria. These pairs feature distinct Aligned (Positive) and Misaligned (Negative) examples suitable for robust preference optimization.

\subsection{Extended Visualization of R$^3$-Bench}
\label{subsec:r3_vis}

In this section, we present additional visualizations to illustrate the diversity of our benchmark and provide a qualitative comparison of R$^3$-Refiner against SOTA UMMs, MLLMs, and reflective visual generation methods.

\begin{figure*}[t]
    \centering
    \includegraphics[width=1.0\linewidth]{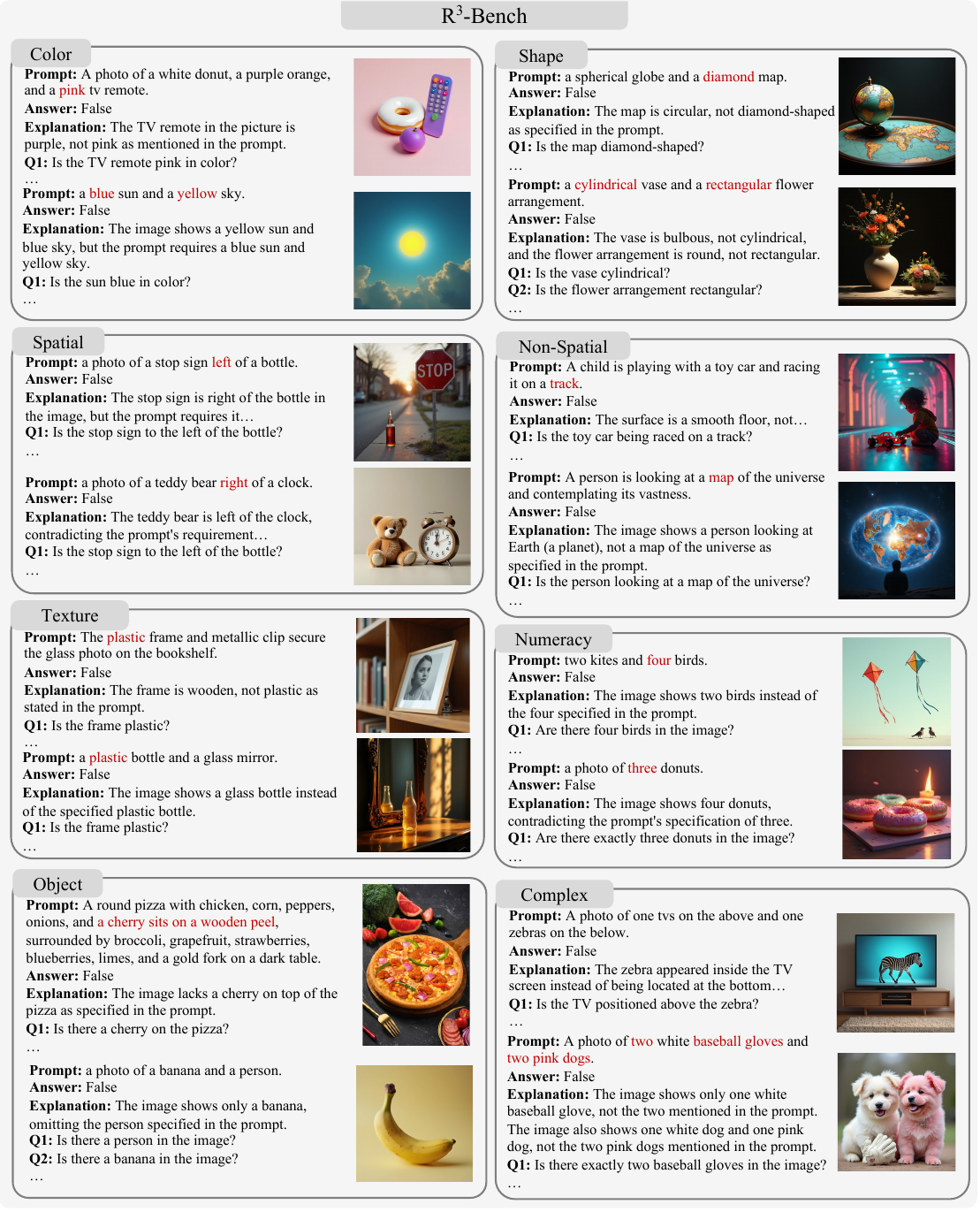}
    \caption{\textbf{Visualizations of R$^3$-Bench.} The benchmark spans eight fine-grained categories (e.g., Spatial, Numeracy, Complex), designed to rigorously test visual reasoning and rectification.}
    \label{fig:r3_bench_examples}
\end{figure*}

\mypara{Visualizations of R$^3$-Bench.}
As illustrated in Fig.~\ref{fig:r3_bench_examples}, R$^3$-Bench is designed to cover a broad spectrum of visual challenges. 
The dataset spans eight fine-grained categories: \textit{Color, Shape, Texture, Spatial, Numeracy, Object, Complex,} and \textit{Non-Spatial}. 
Unlike existing benchmarks that focus on simple object existence, R$^3$-Bench includes ``hard negatives" constructed via our counterfactual rewriting and visual inversion pipelines. 
For instance, the \textit{Spatial} examples require precise understanding of relative positioning (e.g., ``left of vs. right of"), while the \textit{Numeracy} samples demand exact counting in cluttered scenes. 
This diversity ensures that R$^3$-Bench serves as a rigorous testbed for the complete Reason-Reflect-Rectify loop.

\mypara{Qualitative Comparison with SOTA Methods.}
We provide a qualitative comparison between R$^3$-Refiner and varying baselines. As shown in the following figures, R$^3$-Refiner demonstrates superior capability across all three stages of the R$^3$ loop, effectively addressing common failure modes observed in existing methods.

\begin{figure*}[t]
    \centering
    \includegraphics[width=1.0\linewidth]{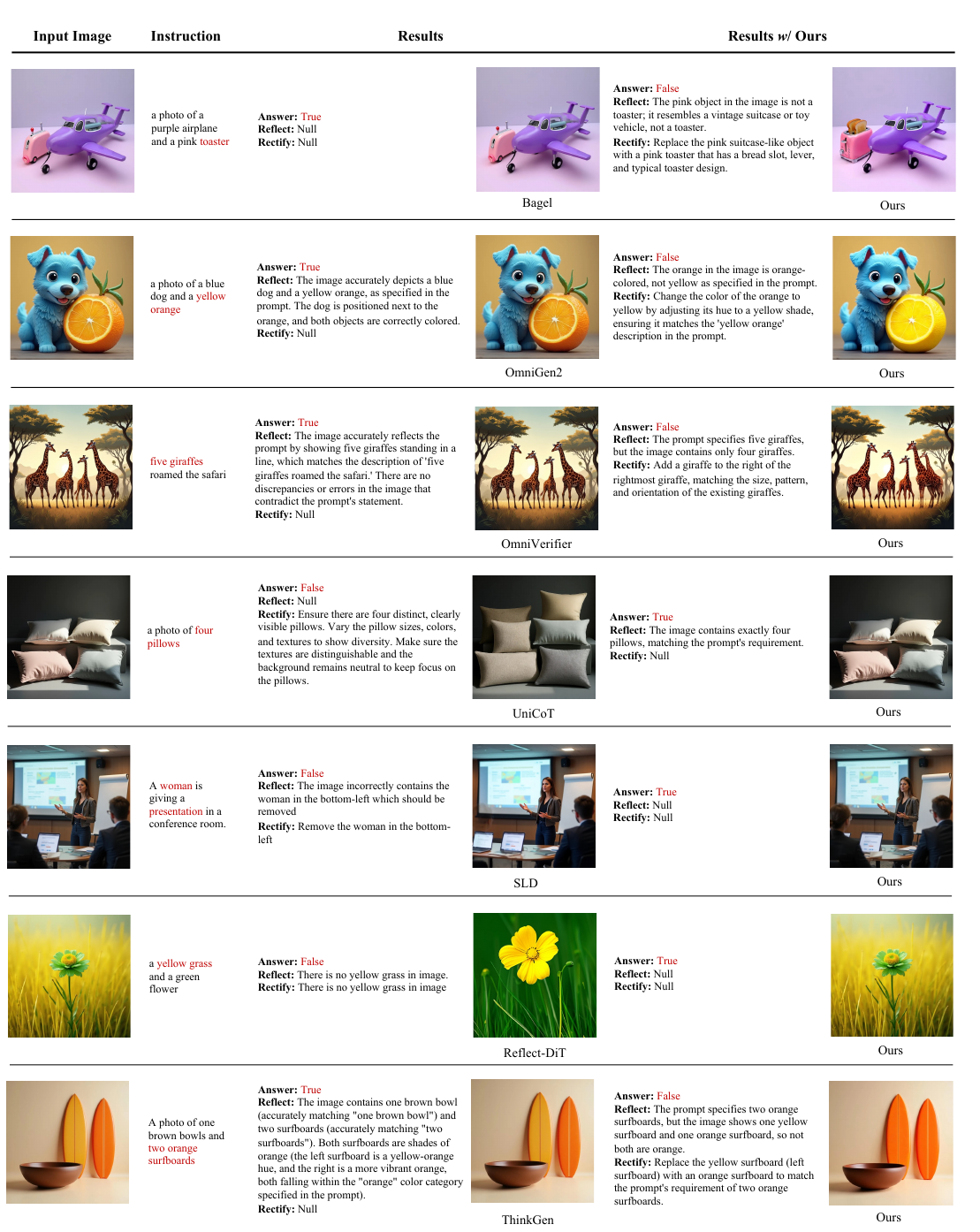}
    \caption{\textbf{Qualitative comparison on Stage I (Verdict).} Baselines like Bagel and OmniGen2 often fail to detect semantic mismatches (e.g., identifying a pink toy car as a ``pink toaster''). R$^3$-Refiner correctly issues a ``False'' verdict based on precise visual evidence.}
    \label{fig:cmp_type_1}
\end{figure*}

\textit{Type I: Verification Failures (Verdict Errors).}
Fig.~\ref{fig:cmp_type_1} illustrates the verdict stage. Baseline models often struggle with fine-grained visual discrimination. For instance, Bagel and ThinkGen frequently output incorrect ``True'' verdicts for mismatched images (e.g., missing objects or wrong colors), exhibiting a strong ``yes-man'' bias. Conversely, some methods like Reflect-DiT may hallucinate errors (False Negatives). R$^3$-Refiner accurately detects these subtle discrepancies, serving as a reliable gatekeeper.

\begin{figure*}[t]
    \centering
    \includegraphics[width=1.0\linewidth]{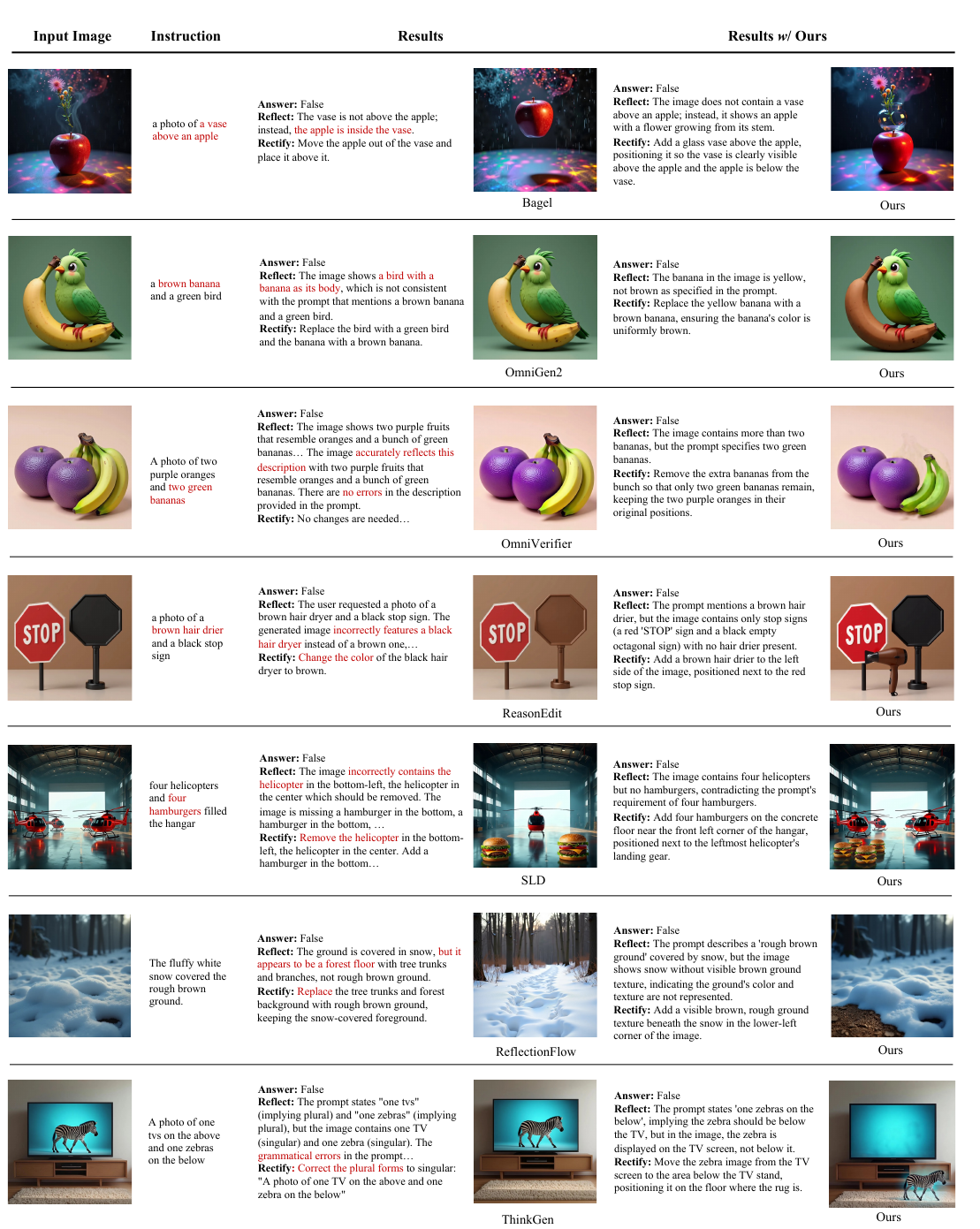}
    \caption{\textbf{Qualitative comparison on Stage II (Reflection).} Existing methods frequently hallucinate details in their explanations. For instance, \textbf{ReasonEdit} attempts to correct the color of a non-existent object. R$^3$-Refiner correctly identifies the root cause (e.g., missing object) without fabrication.}
    \label{fig:cmp_type_2}
\end{figure*}

\textit{Type II: Hallucinated Reflections.}
Fig.~\ref{fig:cmp_type_2} highlights comparisons in the reasoning/explanation stage. Even when baselines correctly identify an image as ``False'', their reasoning is often ungrounded. For example, ReasonEdit criticizes a specific object's color (e.g., ``the hair dryer is black'') even when the object is entirely missing from the image. R$^3$-Refiner avoids such hallucinations, providing explanations that strictly adhere to the visible pixel content.

\begin{figure*}[t]
    \centering
    \includegraphics[width=1.0\linewidth]{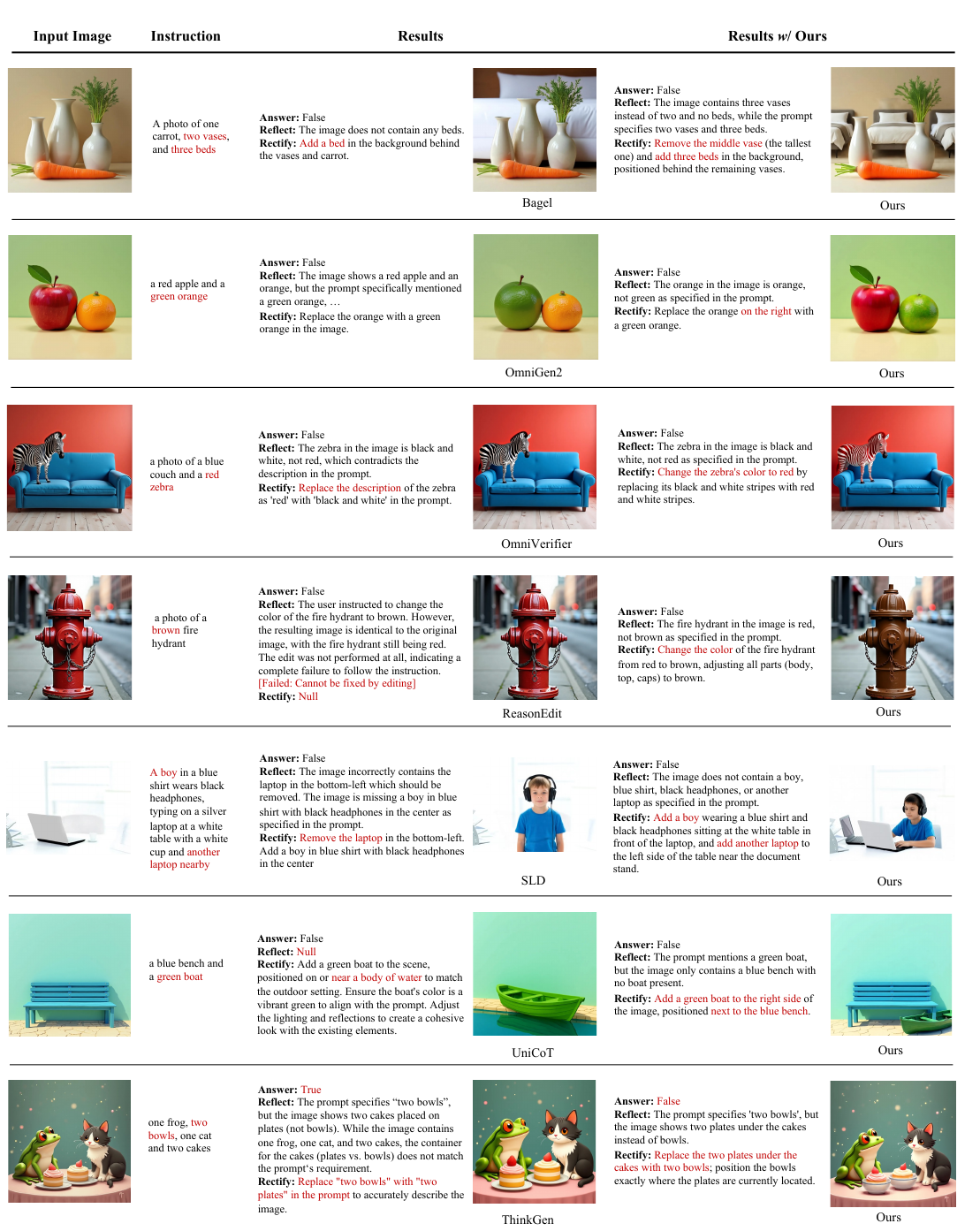}
    \caption{\textbf{Qualitative comparison on Stage III (Rectification).} A common failure mode in baselines (e.g., ThinkGen, OmniVerifier) is proposing to edit the \textit{text prompt} instead of the image. R$^3$-Refiner generates specific image editing instructions (e.g., ``Add a green boat'') to align the visual content with the original prompt.}
    \label{fig:cmp_type_3}
\end{figure*}

\textit{Type III: Evasive vs. Constructive Rectification.}
Fig.~\ref{fig:cmp_type_3} reveals a critical gap in the rectification stage. A pervasive issue with methods like OmniVerifier and ThinkGen is \textit{Evasive Rectification}—they suggest modifying the user's text prompt to match the erroneous image (e.g., ``Replace two bowls with two plates in the prompt'') rather than fixing the image itself. R$^3$-Refiner, by contrast, generates constructive, actionable image editing instructions (e.g., ``Replace the plates with bowls''), fulfilling the user's original intent.

\subsection{Failure Case Analysis}
\label{subsec:failure_case}

\begin{figure}[ht]
    \centering
    \includegraphics[width=0.9\linewidth]{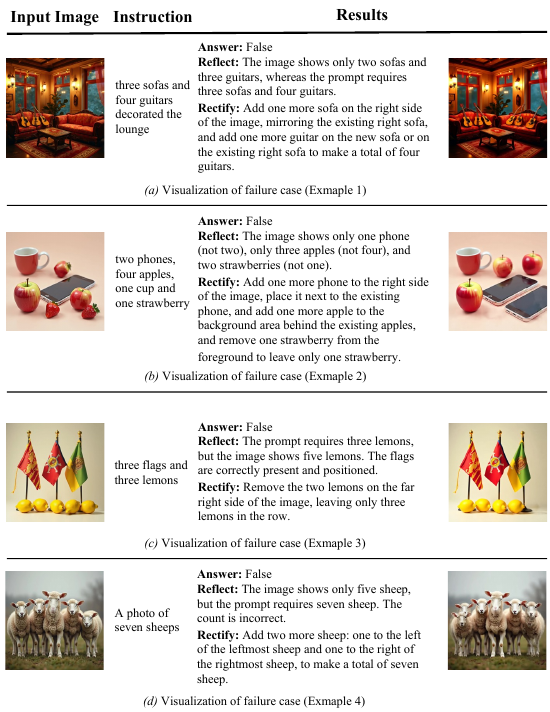}
    \caption{\textbf{Visualization of failure cases.} We provide several visualizations of failure cases, corresponding to two primary failure types: 1) \textbf{Dense
Numeracy Errors} (panels (a) and (b)), 2) \textbf{Editor Capability Limits} (panels (c) and (d)).}
    \label{fig:failure_cases}
\end{figure}

Despite its strong performance, R$^3$-Refiner faces challenges in extreme scenarios. As illustrated in Fig.~\ref{fig:failure_cases}, we identify two primary failure types: (1) Editor Capability Limits, where the policy generates a correct instruction (e.g., ``add a person''), but the backend editor fails to generate a realistic object; and (2) Dense Numeracy Errors, where the model occasionally miscounts objects in highly cluttered scenes (e.g., $>$10 items), likely due to the resolution constraints of the vision encoder.

\section{Additional Quantitative Analysis}
\label{sec:additional_quant_analysis}

\subsection{Evaluator Robustness}
\label{subsec:evaluator_robustness}

To assess the robustness of $\mathcal{S}_{\text{rect}}$ to the choice of automated evaluator, we replace Qwen3-VL-235B~\cite{Qwen3-VL} with GPT-5.2 configured with low reasoning effort and re-run the R$^3$-Bench rectification evaluation without changing any other component of the pipeline. The comparison includes R$^3$-Refiner-BG trained with Bagel~\cite{bagel}, GPT-4o~\cite{gpt4o}, Qwen-family MLLMs~\cite{qwenvl2_5,Qwen3-VL}, and existing verifier-based methods including SLD~\cite{sld} and OmniVerifier~\cite{omniverifier}. As shown in Tab.~\ref{tab:evaluator_robustness}, the two evaluators produce highly consistent model rankings. The only minor discrepancy is between R$^3$-Refiner-BG and GPT-5.2, whose scores are nearly tied under both evaluators.

\begin{table}[tbh]
\centering
\caption{\textbf{Evaluator robustness of $\mathcal{S}_{\text{rect}}$.} We replace Qwen3-VL-235B~\cite{Qwen3-VL} with GPT-5.2 configured with low reasoning effort as the evaluator and report consistent rankings across representative models.}
\label{tab:evaluator_robustness}
\small
\renewcommand{\arraystretch}{1.1}
\setlength{\tabcolsep}{3pt}
\resizebox{\columnwidth}{!}{%
\begin{tabular}{lcccc}
\toprule
\textbf{Model} & \textbf{GPT Eval.} & \textbf{Qwen Eval.} & \textbf{GPT Rank} & \textbf{Qwen Rank} \\
\midrule
R$^3$-Refiner-BG & 0.62 & 0.66 & 1 & 1 \\
GPT-5.2 & 0.62 & 0.65 & 1 & 2 \\
R$^3$-Refiner-QE & 0.58 & 0.62 & 3 & 3 \\
Gemini-3-Pro & 0.55 & 0.60 & 4 & 4 \\
Qwen3-VL-8B & 0.50 & 0.54 & 5 & 5 \\
GPT-4o & 0.49 & 0.53 & 6 & 6 \\
Qwen2.5-VL-7B & 0.36 & 0.38 & 7 & 7 \\
SLD & 0.25 & 0.28 & 8 & 8 \\
OmniVerifier & 0.17 & 0.17 & 9 & 9 \\
\bottomrule
\end{tabular}%
}
\vspace{-0.75em}
\end{table}

\subsection{Training-Editor Transfer}
\label{subsec:cross_editor_generalization}

To examine whether R$^3$-Refiner transfers across training editors, we train two variants with different editors and evaluate them under multiple inference-time editors. R$^3$-Refiner-QE is trained with Qwen-Image-Edit, while R$^3$-Refiner-BG is trained with Bagel. As shown in Tab.~\ref{tab:cross_editor_generalization}, both variants consistently improve over their corresponding baselines across open-source and closed-source editors, supporting training-editor transfer.

\begin{table}[tbh]
\centering
\caption{\textbf{Training-editor transfer.} We train R$^3$-Refiner with different editors and evaluate each variant under multiple inference-time editors on GenEval++ and R$^3$-Bench.}
\label{tab:cross_editor_generalization}
\scriptsize
\renewcommand{\arraystretch}{1.03}
\setlength{\tabcolsep}{0pt}
\begin{tabular*}{\columnwidth}{@{\extracolsep{\fill}}lccc@{}}
\toprule
\multicolumn{4}{@{}l}{\emph{(a) GenEval++ Avg}} \\
\midrule
\textbf{Method} & \textbf{GPT Image} & \textbf{Qwen-Image} & \textbf{OmniGen2} \\
\midrule
Baseline & 0.793 & 0.654 & 0.314 \\
+ R$^3$-Refiner-QE & 0.829 & 0.714 & 0.364 \\
+ R$^3$-Refiner-BG & \textbf{0.839} & 0.711 & 0.343 \\
\addlinespace[2pt]
\midrule
\multicolumn{4}{@{}l}{\emph{(b) R$^3$-Bench $\mathcal{S}_{\text{rect}}$}} \\
\midrule
\textbf{Method} & \textbf{Qwen-Edit} & \textbf{Banana} & \textbf{GPT Image} \\
\midrule
GPT-5.2 & 0.65 & 0.55 & \textbf{0.72} \\
Gemini-3-Pro & 0.60 & 0.56 & 0.71 \\
Qwen3-VL-8B & 0.54 & 0.49 & 0.63 \\
+ R$^3$-Refiner-QE & 0.62 & \textbf{0.58} & \textbf{0.72} \\
+ R$^3$-Refiner-BG & \textbf{0.66} & \textbf{0.58} & 0.71 \\
\bottomrule
\end{tabular*}
\vspace{-0.75em}
\end{table}

\subsection{Benchmark Reliability}
\label{subsec:benchmark_reliability}

R$^3$-Bench is designed to evaluate whether models can diagnose semantically verifiable compositional errors and translate these diagnoses into effective rectification actions. It is not intended to exhaustively cover all generation failures. Within this scope, we curate 670 expert-annotated test samples to enable controlled factual VQA evaluation while keeping human verification cost manageable. We assess whether this scale yields reliable and discriminative model comparisons through two complementary analyses.

\mypara{Paired Bootstrap.}
First, we run paired bootstrap over the 670 test samples. We resample them with replacement for $B=1000$ rounds using shared indices across models and seed 42, and compute 95\% confidence intervals for pairwise $\mathcal{S}_{\text{rect}}$ differences between R$^3$-Refiner-BG and representative baselines. As shown in Tab.~\ref{tab:benchmark_reliability}(a), R$^3$-Refiner-BG is statistically distinguishable from Gemini-3-Pro, Qwen3-VL-8B, and OmniVerifier at $\alpha=0.05$, while its comparison with GPT-5.2 is non-significant. Given their small $\mathcal{S}_{\text{rect}}$ gap, this result is more consistent with a near-tie than with benchmark instability.

\mypara{Rank Stability.}
We further evaluate rank stability by drawing stratified subsamples for 500 rounds at each sample size and computing Kendall's $\tau$ against the full-set ranking. As shown in Tab.~\ref{tab:benchmark_reliability}(b), the ranking remains stable under subsampling, reaching $\tau=0.95$ at $n=400$. When the non-significant R$^3$-Refiner-BG/GPT-5.2 pair is treated as tied, the subsampled ranking exactly matches the full-set ranking at $n=400$.

\begin{table}[tbh]
\centering
\caption{\textbf{Benchmark reliability analysis.} Paired bootstrap CIs and rank stability under subsampling. $\Delta\mathcal{S}_{\text{rect}}$ is computed as R$^3$-Refiner-BG minus the compared model.}
\label{tab:benchmark_reliability}
\scriptsize
\renewcommand{\arraystretch}{1.03}
\setlength{\tabcolsep}{0pt}
\begin{tabular*}{\columnwidth}{@{\extracolsep{\fill}}lcccc@{}}
\toprule
\multicolumn{5}{@{}l}{\emph{(a) Paired bootstrap CIs}} \\
\midrule
\textbf{Model} & \textbf{$\mathcal{S}_{\text{rect}}$} & \textbf{CI Low} & \textbf{CI High} & \textbf{Sig.} \\
\midrule
GPT-5.2 & 0.653 & $-0.040$ & $+0.050$ & No \\
Gemini-3-Pro & 0.599 & $+0.008$ & $+0.109$ & Yes \\
Qwen3-VL-8B & 0.544 & $+0.064$ & $+0.163$ & Yes \\
OmniVerifier & 0.167 & $+0.427$ & $+0.555$ & Yes \\
\end{tabular*}

\vspace{0.35em}
\begin{tabular*}{\columnwidth}{@{\extracolsep{\fill}}lccc@{}}
\toprule
\multicolumn{4}{@{}l}{\emph{(b) Rank stability}} \\
\midrule
\textbf{Metric} & \textbf{$n=200$} & \textbf{$n=400$} & \textbf{Full} \\
\midrule
Kendall's $\tau$ & 0.92 & 0.95 & 1.00 \\
Kendall's $\tau^\dagger$ & 0.97 & 1.00 & 1.00 \\
Exact Match & 87\% & 100\% & 100\% \\
\bottomrule
\end{tabular*}
\vspace{-0.75em}
\end{table}

\subsection{Iterative Refinement Analysis}
\label{sec:iterative_refinement_analysis}

\begin{figure}[h]
    \centering
    \includegraphics[width=0.9\linewidth]{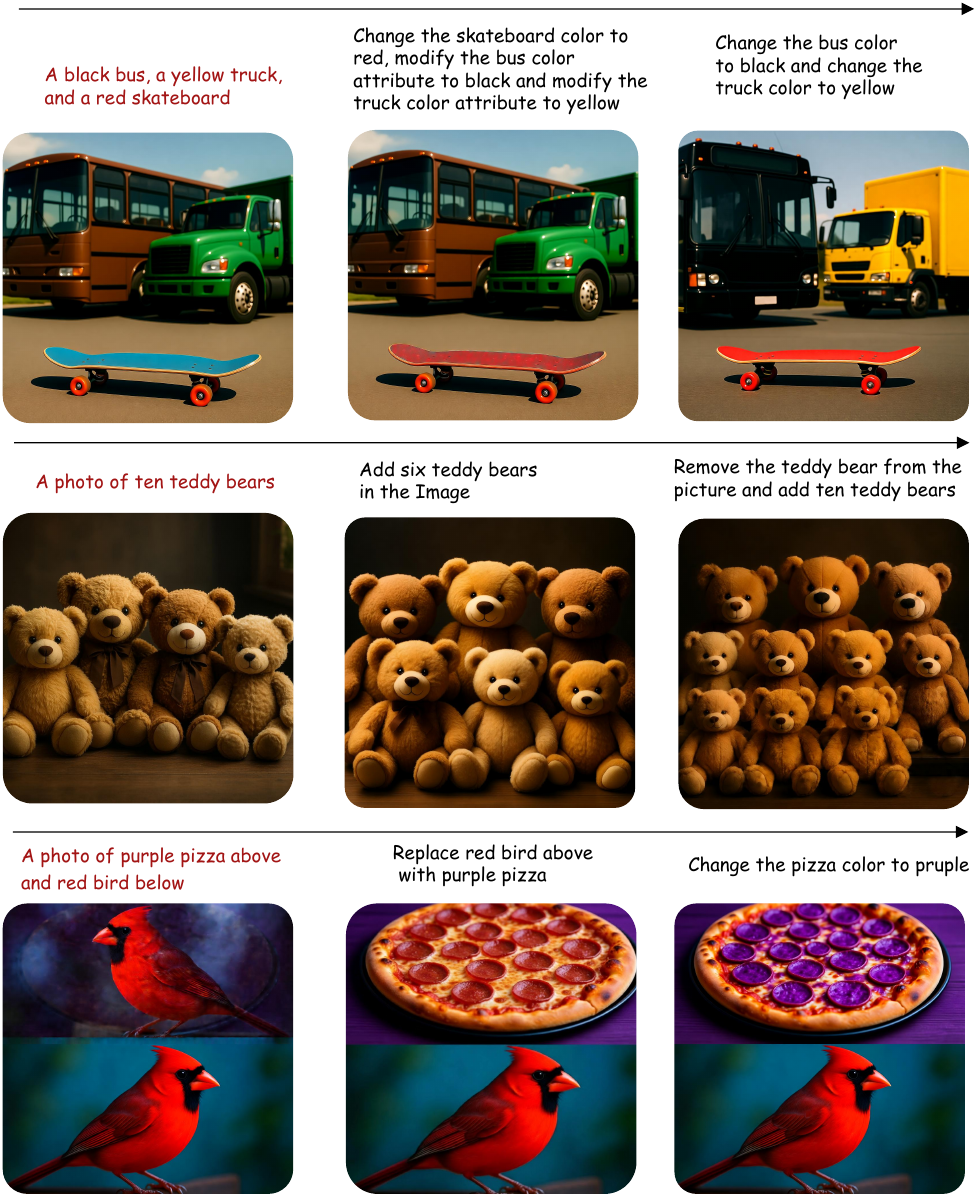}
    \caption{\textbf{Qualitative visualization of the iterative refinement process.}}
    \label{fig:inference_scaling_visual}
\end{figure}

In this section, we explicitly analyze the iterative rectification capability of R$^3$-Refiner through a representative qualitative case study. 
As illustrated in Fig.~\ref{fig:inference_scaling_visual}, the refinement trajectory is visualized in three stages: the Left panel displays the initial image generated by the base model, which contains visual inconsistencies; the Middle panel presents the improved result after the first round of modification; and the Right panel shows the final output after the second round of modification, achieving full alignment with the target prompt.

\section{Training Implementation Details}
\label{sec:train_details}

\subsection{Implementation Hyperparameters}
\label{subsec:hyperparams}

We utilize Qwen2.5-VL-7B-Instruct and Qwen3-VL-8B-Instruct as our base policy models $\pi_\theta$. The edit model is Qwen-Image-Edit-2511. The optimization is performed using the Group Relative Policy Optimization (GRPO) algorithm driven by the Hierarchical Reward Mechanism (HRM) defined in Sec.~\ref{subsec:r3_refiner}. The full training process takes approximately 3 days.

Tab.~\ref{tab:hyperparameters} lists the detailed hyperparameters. Note that the Stage weights ($\alpha_1, \alpha_2$) act as global scaling factors balancing the reasoning phase ($R_{\text{reason}}$) and rectification phase ($R_{\text{rect}}$). Within Stage I, the Accuracy Weight ($\lambda_{\text{acc}}$) and Format Weights ($\lambda_{\text{fmt}}$) specifically govern the trade-off between verdict correctness and structural compliance.

\begin{table}[t]
    \centering
    \caption{Detailed hyperparameter settings of R$^3$-Refiner preference optimization training.}
    \label{tab:hyperparameters}
    \small
    \renewcommand{\arraystretch}{1.25}
    \setlength{\tabcolsep}{3pt}
    \resizebox{\columnwidth}{!}{%
    \begin{tabular}{lcl}
        \toprule
        \textbf{Hyperparameter} & \textbf{Value} & \textbf{Description} \\
        
        \midrule
        \multicolumn{3}{l}{\textit{General Optimization}} \\
        Optimizer & AdamW & With $\beta_1\!=\!0.9, \beta_2\!=\!0.999$. \\
        Learning Rate & $1 \times 10^{-6}$ & With cosine decay scheduler. \\
        Weight Decay & $1 \times 10^{-2}$ & L2 regularization coefficient. \\
        Global Batch Size & 128 & Total batch size per update step. \\
        Micro Batch Size & 4 & Per-device batch size for gradient accumulation. \\
        Epochs & 5 & Total training epochs. \\
        Max Prompt Length & 2560 & Maximum input tokens including image tokens. \\
        Max Response Length & 2048 & Maximum generated output tokens. \\
        
        \midrule
        \multicolumn{3}{l}{\textit{GRPO Algorithm}} \\
        Advantage Estimator & GRPO & Group Relative Policy Optimization. \\
        Group Size ($N$) & 8 & Rollout samples per prompt for advantage estimation. \\
        KL Coefficient ($\lambda_{\text{kl}}$) & $1 \times 10^{-2}$ & Weight for KL divergence penalty. \\
        Clip Ratio & $[0.2, 0.28]$ & Asymmetric PPO clipping range. \\
        
        \midrule
        \multicolumn{3}{l}{\textit{Hierarchical Self-Rectification Rewards}} \\
        Stage-1 Weight ($\alpha_1$) & 0.25 & Weight for initial verification reward. \\
        Stage-2 Weight ($\alpha_2$) & 0.75 & Weight for post-rectification reward. \\
        Accuracy Weight ($\lambda_{\text{acc}}$) & 0.7 & Base reward for correct verification verdict. \\
        Think Format Weight & 0.1 & Penalty for invalid thinking format ($\lambda_{\text{fmt}}$). \\
        JSON Format Weight & 0.2 & Penalty for invalid JSON format ($\lambda_{\text{fmt}}$). \\
        
        \midrule
        \multicolumn{3}{l}{\textit{Sampling Configuration}} \\
        Temperature (Train) & 1.0 & Exploration temperature during rollout. \\
        Temperature (Eval) & 0.01 & Near-greedy decoding for evaluation. \\
        Top-$p$ (Train) & 1.0 & No nucleus sampling truncation. \\
        Top-$p$ (Eval) & 0.001 & Near-deterministic decoding. \\
        
        \midrule
        \multicolumn{3}{l}{\textit{Data \& Image Processing}} \\
        Rollout Batch Size & 128 & Batch size for generating rollouts. \\
        Min Pixels & $512^2$ & Minimum image resolution. \\
        Max Pixels & ${\sim}1088^2$ & Maximum image resolution. \\
        
        \midrule
        \multicolumn{3}{l}{\textit{Infrastructure}} \\
        Training Time & $\sim$3 days & Total wall-clock training duration. \\
        
        \bottomrule
    \end{tabular}%
    }
\end{table}

\section{Evaluation Metrics Details}
\label{sec:metrics}

To comprehensively assess the performance of the R$^3$ pipeline, we introduce specific metrics aligned with the two-phase protocol defined in Sec.~\ref{subsec:evaluation_protocol}: Verdict-Reflection Alignment (Phase I) and Rectification Efficacy (Phase II). These metrics provide a rigorous evaluation by explicitly validating the correctness of the underlying reasoning process and quantifying the effective visual improvement relative to the error space.

\subsection{Phase I: Reflective Verdict Score ($\mathcal{S}_{\text{ref}}$)}
\label{subsec:reflect_score}

The Reflective Verdict Score evaluates the fidelity of the model's diagnostic capability. Unlike simple binary classification metrics, $\mathcal{S}_{\text{ref}}$ imposes a strictly unified standard that penalizes ``correct guesses'' lacking valid reasoning.

\mypara{Metric Formulation.}
The score $s_i$ for a single sample is calculated based on the ground truth verdict $v_i$.

\textit{For Aligned Samples ($v_i = \text{True}$).}
Since the image matches the prompt, no error explanation is required. The metric degrades to a rule-based binary check:
\begin{equation*}
    s_i = \mathbb{I}(\hat{v}_i = \text{True})
\end{equation*}

\textit{For Misaligned Samples ($v_i = \text{False}$).}
This is the critical evaluation scenario. Correctness requires the model to satisfy two conditions simultaneously: \textit{verdict correctness}, where the model must correctly identify the mismatch ($\hat{v}_i = \text{False}$), and \textit{reasoning validity}, where the model's explanation $\hat{e}_i$ must be semantically equivalent to the ground truth diagnosis $e_i$. We verify the second condition using an LLM-Judge function $\mathcal{J}(e_i, \hat{e}_i)$ (see system prompt in Fig.~\ref{fig:eval_prompt_explanation}, Appendix~\ref{subsec:eval_prompts}). Thus:
\begin{equation*}
    s_i = \mathbb{I}(\hat{v}_i = \text{False}) \cdot \mathcal{J}(e_i, \hat{e}_i)
\end{equation*}

\mypara{Design Rationale.}
This unified metric ensures that the model is not merely guessing the label but possesses a true comprehension of the visual discrepancies. By requiring explanation consistency for negative samples, we filter out spurious correctness.

\subsection{Phase II: Rectification Score ($\mathcal{S}_{\text{rect}}$)}
\label{subsec:rectify_score}

The Rectification Score assesses the ``action efficacy'' of the model, specifically measuring the net gain in visual alignment after editing. We adopt a normalized formulation to rigorously quantify how much of the problem was solved.

\mypara{Metric Formulation.}
We employ a VQA-based alignment function $\mathcal{V}(I, Q) \in [0, 1]$, which aggregates the verification results of atomic questions $Q_i$ decomposed from the prompt. The process involves three sequential steps:

\textit{Decomposition.} The prompt $P_i$ is decomposed into atomic boolean questions $Q_i$ (see decomposition prompt in Fig.~\ref{fig:prompt_decomposition}, Appendix~\ref{subsec:eval_prompts}).

\textit{Evaluation.} We calculate the alignment scores for both the original misaligned image ($S_{\text{pre}} = \mathcal{V}(I_i^{(t)}, Q_i)$) and the rectified image ($S_{\text{post}} = \mathcal{V}(I_i^{(t+1)}, Q_i)$) using the VQA verification prompt (see Fig.~\ref{fig:eval_prompt_vqa}, Appendix~\ref{subsec:eval_prompts}).

\textit{Normalization.} Finally, the score represents the gain ($S_{\text{post}} - S_{\text{pre}}$) normalized by the maximum possible gain ($1 - S_{\text{pre}}$):
\begin{equation}
    \mathcal{S}_{\text{rect}} = \frac{1}{N_{\texttt{neg}}} \sum_{i: v_i = \text{False}} \frac{\mathcal{V}(I_i^{(t+1)}, Q_i) - \mathcal{V}(I_i^{(t)}, Q_i)}{1 - \mathcal{V}(I_i^{(t)}, Q_i)}
\end{equation}

\mypara{Design Rationale.}
A ``misaligned'' input image is rarely 100\% incorrect; it often partially matches the prompt (e.g., correct object but wrong color). Therefore, simply scoring the absolute quality of the final image is insufficient. We focus on measuring the \textit{relative improvement}—the proportion of the previously unresolved error space that is successfully bridged by the model.

\mypara{Metric Interpretation.}
The $\mathcal{S}_{\text{rect}}$ provides a distinct physical meaning regarding the editing quality: a score $> 0$ indicates valid visual improvement where the model successfully fixed errors; a score $\approx 1$ implies the error was completely resolved; conversely, a score $\leq 0$ denotes ineffective editing or degradation where the process introduced new errors.


\section{Prompt Details}
\label{sec:prompts}

In this section, we provide the exact prompt templates used in our R$^3$-Refiner framework and the baseline comparisons.

\tcbset{
    promptbox/.style={
        colback=gray!5, 
        colframe=gray!60, 
        fonttitle=\bfseries, 
        coltitle=black, 
        boxrule=1pt,
        sharp corners,
        rounded corners=southeast,
        arc=4pt,
        left=6pt, right=6pt, top=6pt, bottom=6pt
    }
}

\subsection{Training Prompt for R$^3$-Refiner}
\label{subsec:training_prompt}

Fig.~\ref{fig:prompt_r3} details the instruction employed by our policy $\pi_\theta$, designed to elicit the complete R$^3$ loop defined in Sec.~\ref{subsec:task_formalization}.
To facilitate complex reasoning, the prompt first requires the model to generate an internal chain-of-thought explicitly encapsulated within \texttt{<think>} tags.
Subsequently, the model outputs the structured tuple $\langle v_t, e_t, a_t \rangle$ in a strict JSON format, where the components correspond to the \texttt{"answer"} (verification), \texttt{"explanation"} (reflection), and \texttt{"edit\_prompt"} (rectification) fields, respectively.

\begin{figure*}[t]
    \centering
    \small
    
    \begin{tcolorbox}[
        colback=gray!5!white,
        colframe=gray!75!black,
        title=\textbf{Training Prompt for R$^3$-Refiner},
        fonttitle=\bfseries,
        boxrule=0.8pt,
        arc=2mm,
        left=6pt, right=6pt, top=6pt, bottom=6pt
    ]

    \textbf{User:} \textcolor{blue}{\texttt{<image>}} This image was generated from the prompt: $\textcolor{blue}{\{\{ \textnormal{content} \mid \textnormal{trim} \}\}}$. Please carefully analyze the image and determine whether all the objects, attributes, count, and spatial relationships mentioned in the prompt are correctly represented in the image.
    
    \vspace{1em}
    If the image accurately reflects the prompt, please answer ``true"; otherwise, answer ``false".
    
    \vspace{1em}
    \textbf{When the answer is false, you must:}
    \begin{enumerate}[leftmargin=1.5em, nosep, label=\arabic*.]
        \item Identify the main error(s) and describe them briefly in ``explanation".
        \begin{itemize}[leftmargin=1em, label=$\bullet$, nosep]
            \item Clearly state what the prompt requires vs. what is actually shown in the image.
            \item If there are multiple discrepancies (e.g., object missing, wrong color, wrong position, wrong count), you should mention all of the important ones in a concise way.
        \end{itemize}
        \vspace{0.3em}
        \item In ``edit\_prompt", provide a \textbf{direct and specific image editing instruction} to fix the error.
        \begin{itemize}[leftmargin=1em, label=$\bullet$, nosep]
            \item Choose the most appropriate action based on the actual error: add / remove / replace / move / change color / change shape / change texture / modify attribute / adjust count / resize / swap positions
            \item The instruction can contain multiple coordinated edits in one sentence (e.g., change color + add object + move object), as long as it is clear and executable.
            \item The instruction must specify the exact action and the location or reference point when relevant (e.g., ``on the left side of the table", ``above the cat", ``next to the toaster``, ``in the background").
           \item If the target position has no space, propose an alternative that is executable (e.g., swap positions, move the other object, resize first).
        \end{itemize}
    \end{enumerate}

    \vspace{1em}
    \textbf{Examples of good edit\_prompt instructions:}
    \begin{itemize}[leftmargin=1.5em, label=$\bullet$, nosep]
        \item ``Replace the white candle on the right side of the image with a white candle holder."
        \item ``Change the fork's color from silver to gold in the image."
        \item ``Remove the single baseball glove from the left and add a kite in the sky on the left, and remove the kites from the right and add three baseball gloves to the ground on the right."
        \item ``Move the cat to the left side of the pizza so that the cat is clearly positioned to the left of the pizza."
        \item ``Add one more donut to the plate on the left side so that there are exactly three donuts."
        \item ``Swap the positions of the giraffe and the traffic light so that the giraffe is clearly to the right of the traffic light, keeping both fully visible."
    \end{itemize}

    \vspace{1em}
    Respond strictly in the following JSON format:
    \lstset{
        basicstyle=\ttfamily\footnotesize,
        breaklines=true,
        breakindent=10pt,
        breakatwhitespace=true,
        columns=fullflexible,
        keepspaces=true
    }
    \begin{lstlisting}
{
    "answer": true/false,
    "explanation": "A brief, specific description of the main error (if answer is false).",
    "edit_prompt": "A concrete, location-specific editing instruction to fix the error (if answer is false)."
}
    \end{lstlisting}
    You should first think about the reasoning process in the mind and then provide the user with the answer. The reasoning process is enclosed within \texttt{<think>} ... \texttt{</think>} tags, i.e. \texttt{<think>} reasoning process here \texttt{</think>}answer here
    \end{tcolorbox}
\caption{The prompt used to train the R$^3$-Refiner policy. 
This prompt enforces the iterative R$^3$ loop (Sec.~\ref{subsec:task_formalization}) by requiring the model to first generate an explicit internal reasoning process (CoT) before producing the structured tuple $\langle v_t, e_t, a_t \rangle$. 
These components are mapped to the JSON fields \texttt{"answer"} (corresponding to $v_t$, \textit{Reason}), \texttt{"explanation"} (corresponding to $e_t$, \textit{Reflect}), and \texttt{"edit\_prompt"} (corresponding to $a_t$, \textit{Rectify}), ensuring precise alignment with the formalized task definition.}
    \label{fig:prompt_r3}
\vspace{15em}
\end{figure*}

\subsection{Evaluation Prompts}
\label{subsec:eval_prompts}

To ensure reproducibility, we provide the exact prompts used for the LLM-Judge ($\mathcal{J}$) in Phase I and the VQA-based alignment function ($\mathcal{V}$) in Phase II.

\mypara{Phase I: Verdict-Reflection Alignment.}
\label{mypara:eval_phase1}
Fig.~\ref{fig:eval_prompt_explanation} presents the system prompt used by the external LLM-Judge $\mathcal{J}$. This prompt is designed to evaluate the semantic equivalence between the generated reflection $\hat{e}_i$ and the ground truth explanation $e_i$, which is the core component of the Reflective Verdict Score ($\mathcal{S}_{\text{ref}}$).

\begin{figure*}[h!]
    \centering
   \small
    \begin{tcolorbox}[
        colback=gray!5!white,
        colframe=gray!75!black,
        title=\textbf{Phase I Evaluation: LLM-Judge for Verdict-Reflection Alignment ($\mathcal{J}$)},
        fonttitle=\bfseries,
        boxrule=0.8pt,
        arc=2mm,
        left=6pt, right=6pt, top=6pt, bottom=6pt
    ]
    \textbf{[System Prompt]}
    
    You are an expert evaluator for image reflection tasks. Your task is to compare two explanations for why an image fails to match a prompt and determine if they are semantically equivalent.
    \vspace{0.5em}

    You are given:
    \begin{itemize}[leftmargin=1.2em, topsep=0pt, itemsep=0pt]
        \item The original image prompt.
        \item A Model Explanation and a GT (Ground Truth) Explanation.
    \end{itemize}
    \vspace{0.5em}
    Typical error dimensions include:
    \begin{itemize}[leftmargin=1.2em, topsep=0pt, itemsep=0pt]
        \item \textbf{Color}: wrong or missing colors.
        \item \textbf{Object}: wrong, missing, or extra objects.
        \item \textbf{Numeracy}: wrong object counts or quantities.
        \item \textbf{Spatial}: wrong positions, relative locations, or spatial relations.
        \item \textbf{Shape}: wrong shapes or geometric properties.
        \item \textbf{Texture}: wrong material or surface appearance.
        \item \textbf{Complex}: complex combinations of multiple basic errors (for example, several objects and relations are all wrong at the same time, or multiple dimensions are intertwined).
        \item \textbf{Non}: more subjective or high-level mismatches that are not purely low-level visual attributes, such as incorrect actions or activities, scene type, atmosphere, style, or other semantic aspects that do not clearly fall into the categories above.
    \end{itemize}
    \vspace{0.5em}
    
    These are general categories that describe how an image can fail to match a prompt (for example: wrong color, wrong object, wrong count, wrong spatial relation, wrong shape or texture, wrong action or atmosphere, etc.).
    \vspace{0.5em}

    \textbf{Definitions:}
    \begin{itemize}[leftmargin=1.2em, topsep=0pt, itemsep=0pt]
        \item A ``core error" is the main reason why the image does NOT satisfy the prompt (for example: wrong object count, wrong object type, wrong attribute, wrong spatial relation, missing or extra object, wrong action, wrong style, etc.).
        \item There can be multiple low-level details, but usually only a small number of core errors.
    \end{itemize}
    \vspace{0.5em}

    The model's explanation is considered correct if it identifies the SAME CORE ERROR as the GT (Ground Truth) explanation, even if:
    \begin{itemize}[leftmargin=1.2em, topsep=0pt, itemsep=0pt]
        \item The wording is different
        \item Additional context or details are mentioned (for example, mentioning other objects that are present)
        \item The phrasing or style differs
    \end{itemize}
    \vspace{0.5em}

    \textbf{IMPORTANT:}
    \begin{itemize}[leftmargin=1.2em, topsep=0pt, itemsep=0pt]
        \item Use the original prompt to understand what the image is supposed to contain or look like.
        \item Focus on whether both explanations point to the SAME fundamental problem in how the image fails to match the prompt, considering the typical dimensions listed above.
        \item Do NOT reject explanations just because one includes extra information or uses different words to describe the same error.
        \item However, if the Model Explanation introduces a NEW, SEPARATE core error that is not implied by the GT Explanation, or criticizes something that is actually correct according to the original prompt, then they are NOT semantically equivalent.
    \end{itemize}
    \vspace{0.5em}

    You should respond in JSON format:
    \begin{lstlisting}[
        breaklines=true,
        basicstyle=\ttfamily\small,
        frame=single,
        framerule=0pt,
        xleftmargin=1em,
        xrightmargin=1em
    ]
{
    "is_correct": true/false,
    "reasoning": "A brief explanation of why the explanations are or are not semantically equivalent."
}
    \end{lstlisting}

    \vspace{0.5em}
    \hrule
    \vspace{0.5em}
    
    \textbf{[User Prompt]}
    
    Original Prompt:
    $\textcolor{blue}{\{\textnormal{original\_prompt}\}}$

    Compare the following two explanations:
    \begin{itemize}[leftmargin=1.2em, topsep=0pt, itemsep=0pt]
        \item Model Explanation: $\textcolor{blue}{\{\textnormal{model\_explanation}\}}$
        \item  GT Explanation: $\textcolor{blue}{\{\textnormal{gt\_explanation}\}}$
    \end{itemize}

    Are these two explanations semantically equivalent? Respond in JSON format as specified.
    \end{tcolorbox}
    \caption{The exact system prompt used by the LLM-Judge $\mathcal{J}$ in Phase I. This judge evaluates whether the generated reflection $\hat{e}_i$ is semantically equivalent to the ground truth explanation $e_i$, which is used to compute the Reflective Verdict Score ($\mathcal{S}_{\text{ref}}$).}
    \label{fig:eval_prompt_explanation}
\end{figure*}

\mypara{Phase II: Rectification Efficacy.}
\label{mypara:eval_phase2}
This phase quantifies the improvement of the rectified image using the Rectification Score ($\mathcal{S}_{\text{rect}}$). This process involves two steps: (1) decomposing the prompt into atomic questions, and (2) verifying these questions against the image.

\textit{Question Decomposition:} 
To support fine-grained evaluation, we decompose the target prompt $P_i$ into a set of atomic Boolean questions $Q_i$. Fig.~\ref{fig:prompt_decomposition} presents the few-shot prompt used for this decomposition task.

\textit{VQA-based Verification:} 
Fig.~\ref{fig:eval_prompt_vqa} displays the prompt template used for the VQA-based alignment function $\mathcal{V}$. This function applies an external MLLM to answer the decomposed questions $Q_i$, producing the probabilities used to calculate $\mathcal{S}_{\text{rect}}$.

\begin{figure*}[h!]
    \centering
   \small
    \begin{tcolorbox}[
        colback=gray!5!white,
        colframe=gray!75!black,
        title=\textbf{Question Decomposition Prompt (Phase II)},
        fonttitle=\bfseries,
        boxrule=0.8pt,
        arc=2mm,
        left=6pt, right=6pt, top=6pt, bottom=6pt
    ]
    Convert the given prompt into a JSON object containing a list of simple, verifiable boolean questions.
    The questions should focus on the prompt's main requirement, related to categories: Color, Shape, Texture, Spatial, Numeracy, Object, Complex, Non.
    For 'Non', generate questions that verify the main subjects, actions, and scene description.
    \vspace{0.5em}

    \textbf{You MUST:}
    \begin{itemize}[leftmargin=1.2em, topsep=0pt, itemsep=0pt, nosep]
        \item Focus on atomic facts (objects, attributes, relations, actions, counts).
        \item Make each question answerable as a boolean fact.
        \item Do NOT include any answers in the JSON.
    \end{itemize}

    \vspace{0.5em}
    \hrule
    \vspace{0.5em}

    \begin{minipage}[t]{0.49\textwidth}
        \textbf{Example 1 (Numeracy):} \\
        Input: "one rabbit and three horses" (numeracy) \\
        \texttt{\{"yn\_question\_list": ["Is there a rabbit in the image?", "Is there exactly one rabbit?", "Are there horses in the image?", "Are there exactly three horses?"]\}}

        \vspace{0.4em}
        \textbf{Example 2 (Color):} \\
        Input: "a gold bench and a green clock" (color) \\
        \texttt{\{"yn\_question\_list": ["Is there a bench in the image?", "Is the bench gold in color?", "Is there a clock in the image?", "Is the clock green in color?"]\}}

        \vspace{0.4em}
        \textbf{Example 3 (Spatial):} \\
        Input: "A horse on the right of a man" (spatial) \\
        \texttt{\{"yn\_question\_list": ["Is there a horse in the image?", "Is there a man in the image?", "Is the horse to the right of the man?"]\}}

        \vspace{0.4em}
        \textbf{Example 4 (Shape):} \\
        Input: "A circular chandelier..." (shape) \\
        \texttt{\{"yn\_question\_list": ["Is there a chandelier?", "Is the chandelier circular?", "Is there a wall art?", "Is the wall art triangular?"]\}}
    \end{minipage}
    \hfill \vrule \hfill
    \begin{minipage}[t]{0.49\textwidth}
        \textbf{Example 5 (Texture):} \\
        Input: "a plastic bottle and fabric pants" (texture) \\
        \texttt{\{"yn\_question\_list": ["Is there a bottle?", "Is the bottle plastic?", "Is there a pair of pants?", "Are the pants made of fabric?"]\}}

        \vspace{0.4em}
        \textbf{Example 6 (Object):} \\
        Input: "a surfboard and a knife" (object) \\
        \texttt{\{"yn\_question\_list": ["Is there a surfboard?", "Is there a knife in the image?"]\}}

        \vspace{0.4em}
        \textbf{Example 7 (Complex):} \\
        Input: "The sweet chocolate chip..." (complex) \\
        \texttt{\{"yn\_question\_list": ["Is there a cookie?", "Is the cookie sweet?", "Is there a crust?", "Is the crust crunchy?", "Is there ice cream?", "Is the ice cream soft?", "Did the cookie crumble on them?"]\}}

        \vspace{0.4em}
        \textbf{Example 8 (Non):} \\
        Input: "A child is playing..." (non) \\
        \texttt{\{"yn\_question\_list": ["Is there a child?", "Is the child playing with a toy airplane?", "Does the scene depict a child making airplane noises?"]\}}
    \end{minipage}

    \vspace{0.5em}
    \hrule
    \vspace{0.5em}

    Now, perform the conversion for the following prompt: \\
    Input Prompt: $\textcolor{blue}{\{\textnormal{original\_prompt}\}}$ \quad Input Category: $\textcolor{blue}{\{\textnormal{category}\}}$ \\
    Output JSON:
    \end{tcolorbox}
    \caption{The few-shot prompt (8 examples) used to decompose user prompts into atomic boolean questions. To save space, the JSON outputs in the examples are displayed in a compact format; the actual prompt uses standard JSON indentation.}
    \label{fig:prompt_decomposition}
\end{figure*}

\begin{figure*}[h!]
    \centering
    \small
    \begin{tcolorbox}[
        colback=gray!5!white,
        colframe=gray!75!black,
        title=\textbf{Phase II Evaluation: VQA-based Alignment Function ($\mathcal{V}$)},
        fonttitle=\bfseries,
        boxrule=0.8pt,
        arc=2mm,
        left=6pt, right=6pt, top=6pt, bottom=6pt
    ]
    You are tasked with conducting a careful examination of the image. Based on the content of the image, please answer the following yes or no questions:

    Questions:
    $\textcolor{blue}{\{\textnormal{questions}\}}$
    \vspace{0.5em}

    \textbf{Note that:}
    \begin{enumerate}[leftmargin=1.5em, topsep=0pt, itemsep=0pt]
        \item Each answer should be on a separate line, starting with ``yes" or ``no", followed by the reason.
        \item The order of answers must correspond exactly to the order of the questions.
        \item Each question must have only one answer.
        \item Directly return the answers to each question, without any additional content.
        \item Each answer must be on its own line!
        \item Make sure the number of output answers equals the number of questions!
    \end{enumerate}
    \end{tcolorbox}
    \caption{The prompt template used for the VQA-based alignment function $\mathcal{V}$ in Phase II. This prompt directs the external MLLM to answer the visual question set $Q_i$, producing the scores required to calculate the Rectification Score ($\mathcal{S}_{\text{rect}}$).}
    \label{fig:eval_prompt_vqa}
\end{figure*}

\subsection{Data Construction Prompts}
\label{subsec:data_construction_prompts}

To ensure reproducibility of our data synthesis pipeline described in Sec.~\ref{subsec:data_pipeline}, we provide the exact prompts of data construction. We first present the generation prompts for Counterfactual Rewriting (Fig.~\ref{fig:prompt_counterfactual}) and Visual Inversion (Fig.~\ref{fig:prompt_visual_inversion}), followed by the filtering prompts used in Rationale Verification (Fig.~\ref{fig:prompt_consistency_check}).
\lstset{
    basicstyle=\ttfamily\footnotesize,
    breaklines=true,
    breakindent=10pt,
    breakatwhitespace=true,
    columns=fullflexible,
    keepspaces=true
}

\begin{figure*}[h!]
    \centering
    \small
    \begin{tcolorbox}[
        colback=gray!5!white,
        colframe=gray!75!black,
        title=\textbf{Data Construction: Counterfactual Rewriting},
        fonttitle=\bfseries,
        boxrule=0.8pt,
        arc=2mm,
        left=6pt, right=6pt, top=6pt, bottom=6pt
    ]
    \textbf{System Instruction:}
    You are a highly precise image caption editor specialized in creating false captions for visual verification tasks.

    \textbf{Your Task:}
    \begin{itemize}[leftmargin=1.2em, nosep]
        \item Given an original image caption (which correctly matches the image),
        \item you must modify it to create a \textbf{slightly but clearly incorrect} caption that would NOT match the original image anymore.
    \end{itemize}
    \vspace{0.5em}

    \textbf{First, analyze the caption type} (numeracy, color, texture, shape, spatial, object, complex, non).
    \textbf{Then, modify according to the caption type:}
    \begin{enumerate}[leftmargin=1.5em, nosep]
        \item \textbf{For numeracy}: Change the number (e.g., ``four'' $\rightarrow$ ``two'').
        \item \textbf{For color}: Change one or more color attributes.
        \item \textbf{For texture/shape}: Change material or shape attributes.
        \item \textbf{For spatial}: Change spatial relationship (e.g., "above" $\rightarrow$ ``below'').
        \item \textbf{For object}: Add/remove/replace an object.
        \item \textbf{For complex/non}: Change attributes or action/verb context slightly.
    \end{enumerate}
    \vspace{0.5em}

    \textbf{Strong requirements:}
    \begin{enumerate}[leftmargin=1.5em, nosep]
        \item Keep the \textbf{overall scene, entities, and structure} similar. Only change \textbf{1 or 2 local details}.
        \item The change must be \textbf{specific and objectively checkable}.
        \item The modification must be big enough to be false, but small enough to be plausible.
    \end{enumerate}
    \vspace{0.5em}

    \textbf{Output format (JSON):}
    \begin{lstlisting}
{
  "false_prompt": "...",      // the modified caption that is now false
  "change_type": "...",       // e.g., "numeracy", "color"
  "changed_detail": "..."     // a short explanation of what changed
}
    \end{lstlisting}
    \textbf{Examples:}
    \textit{Example (numeracy):} Original: "a photo of four coasters"
    \begin{lstlisting}
{"false_prompt": "a photo of two coasters", "change_type": "numeracy", "changed_detail": "Changed number from four to two."}
    \end{lstlisting}
    \hrule
    \vspace{0.5em}

    \textbf{User Input:} Original caption: $\textcolor{blue}{\{\textnormal{orig\_caption}\}}$
    \end{tcolorbox}
    \caption{The system prompt used to generate fine-grained hard negatives via counterfactual rewriting. The model rewrites a correct caption into a misaligned one by altering specific visual attributes.}
    \label{fig:prompt_counterfactual}
\end{figure*}

\begin{figure*}[h!]
    \centering
    \small
    \begin{tcolorbox}[
        colback=gray!5!white,
        colframe=gray!75!black,
        title=\textbf{Data Construction: Visual Inversion},
        fonttitle=\bfseries,
        boxrule=0.8pt,
        arc=2mm,
        left=6pt, right=6pt, top=6pt, bottom=6pt
    ]
    \textbf{System Instruction:}
    You are an expert visual analyst capable of reverse-engineering image editing instructions.

    \textbf{Input:}
    You are provided with two images:
    1. \textbf{Pre-edit Image} (\textcolor{blue}{\textless{image\_1}\textgreater}): The original, misaligned image.
    2. \textbf{Post-edit Image} (\textcolor{blue}{\textless{image\_2}\textgreater}): The successfully edited image that corrects a visual error.
    \vspace{0.5em}

    \textbf{Your Task:}
    Compare the two images to identify the specific visual attribute that was modified. Based on this visual difference, infer the \textbf{Target Prompt ($P$)} that accurately describes the Post-edit Image.

    \textbf{Critical Rules:}
    \begin{itemize}[leftmargin=1.5em, nosep]
        \item Focus ONLY on the visible content of the Post-edit Image.
        \item The generated prompt should imply the correction (e.g., if a red car became blue, the prompt should explicitly specify "a blue car").
        \item Identify the primary category of the change.
    \end{itemize}
    \vspace{0.5em}

    \textbf{Output Format (JSON):}
    \begin{lstlisting}
{
  "target_prompt": "Concise, factual description of the Post-edit Image",
  "change_category": "color|shape|texture|object|numeracy|spatial|non|complex",
  "confidence": "high"
}
    \end{lstlisting}
    \end{tcolorbox}
    \caption{The VLM prompt used for visual inversion. By comparing the pre-edit and post-edit images, the model infers the target prompt $P$ that aligns with the corrected visual state.}
    \label{fig:prompt_visual_inversion}
\end{figure*}

\begin{figure*}[t]
    \centering
    \small
    \begin{tcolorbox}[
        colback=gray!5!white,
        colframe=gray!75!black,
        title=\textbf{Data Filtering: Rationale Verification},
        fonttitle=\bfseries,
        boxrule=0.8pt,
        arc=2mm,
        left=6pt, right=6pt, top=6pt, bottom=6pt
    ]

    \textbf{Step 1: Proposer Prompt (Initial Diagnosis)}
    \par\noindent\rule{\textwidth}{0.4pt}
    \vspace{0.3em}
    
    \textbf{System Instruction:}
    You should first think about the reasoning process in the mind and then provide the user with the answer. The reasoning process is enclosed within \texttt{<think> ... </think>} tags, i.e., \texttt{<think> reasoning process here </think> answer here}.
    \vspace{0.5em}

    \textbf{User Input:} 
    \textcolor{blue}{\textless{image}\textgreater} This image was generated from the prompt: \textcolor{blue}{\{prompt\}}. 
    Please carefully analyze the image and determine whether all the objects, attributes, and spatial relationships mentioned in the prompt are correctly represented in the image. 

    If the image accurately reflects the prompt, please answer 'true'; otherwise, answer 'false'. Respond strictly in the following JSON format: 
    \begin{lstlisting}
{
    "answer": true/false,
    "explanation": "If the answer is false, briefly summarize the main error."
}
    \end{lstlisting}

    \textbf{Step 2: Verifier Prompt (Auditing)}
    \par\noindent\rule{\textwidth}{0.4pt}
    \vspace{0.3em}
    
    \textbf{System Instruction:}
    You are a principal evaluator reviewing a teacher model's (OmniVerifier) judgment on whether an image matches a text prompt. Your task is to verify if the teacher model's explanation is accurate and consistent with the actual image and prompt.
    \vspace{0.5em}

    \textbf{Your review criteria:}
    \begin{enumerate}[leftmargin=1.5em, nosep]
        \item If the teacher model's answer is ``true'': Verify that the explanation correctly describes why the image matches the prompt, and that the described elements actually exist in the image.
        \item If the teacher model's answer is ``false'': Verify that the explanation correctly identifies the actual problems, and that these problems truly exist in the image.
    \end{enumerate}
    \vspace{0.5em}

    \textbf{Output Format:}
    \begin{lstlisting}
{
    "review_result": "pass" or "fail",
    "reasoning": "brief explanation..."
}
    \end{lstlisting}

    \textbf{User Input:}
    \textcolor{blue}{\textless{image}\textgreater}
    \textbf{Prompt:} \textcolor{blue}{\{prompt\}} \\
    \textbf{Teacher Model's Answer:} \textcolor{blue}{\{model\_answer\}} \\
    \textbf{Teacher Model's Explanation:} \textcolor{blue}{\{explanation\}}
    \vspace{0.5em}

    Please review:
    \begin{itemize}[leftmargin=1.5em, nosep]
        \item If answer is ``true'': Does the explanation correctly describe why the image matches? Do the described elements actually exist?
        \item If answer is ``false'': Does the explanation correctly identify the problems? Do these problems truly exist?
    \end{itemize}
    
    \end{tcolorbox}
    \caption{The prompts used in the Rationale Verification phase. The Proposer first generates a verdict with explicit reasoning (Step 1), and the Verifier audits the factual grounding of that explanation (Step 2) to filter hallucinations.}
    \label{fig:prompt_consistency_check}
    \vspace{16em}
\end{figure*}

\section{Details of Test Set Curation}
\label{sec:test_set_details}

The construction of R$^3$-Bench follows a comprehensive four-stage pipeline designed to ensure high semantic diversity and annotation accuracy.

\mypara{Stage 1: Generative Data Sourcing.}
We initially generate approximately 260,000 images using state-of-the-art text-to-image models~\cite{bagel,qwen_image} based on prompts from T2I-R1~\cite{t2i_r1} and GenEval++~\cite{geneval++}. To efficiently identify valuable samples, we apply the Generative Ranking and Automated Cascaded Filtering pipeline proposed in this paper. This process filters the raw data into aligned and misaligned pairs based on image-text consistency.
After splitting the data into training and testing sets, we obtain an initial set of 1,000 candidate samples from this generative stream.

\mypara{Stage 2: Real-world Data Augmentation.}
To enhance domain diversity, we incorporate real-world image editing data from GEdit~\cite{gedit}. We first select English editing instructions and exclude categories unsuitable for visual reflection tasks (e.g., stylistic or background-only changes). Using the \textit{gemini-2.5-flash-image} model, we generate the corresponding edited target images. To ensure image quality, we employ Qwen-VL to calculate the VIE-score~\cite{gedit} and apply a Best-of-N selection strategy. We then synthesize misaligned samples by pairing the \textit{pre-edit} source image with a caption of the \textit{post-edit} image generated by Qwen-VL. This reverse-engineering approach contributes 300 additional challenging samples to the pool.

\mypara{Stage 3: Automated Annotation.}
For the combined pool of 1,300 candidates, we employ advanced MLLMs to generate the necessary benchmark annotations. We use Qwen-VL to generate detailed explanations describing why the images deviate from the text prompts. Subsequently, we utilize \textit{Qwen3-Next} to generate a set of Visual Question Answering (VQA) questions for each sample. These questions serve as the metric for evaluating the effectiveness of the rectification actions.

\mypara{Stage 4: Human Verification and Refinement.}
To guarantee the gold-standard quality of R$^3$-Bench, human experts perform a final round of strict verification. Experts review the binary consistency labels to correct any automated judgment errors. They also refine the generated explanations for clarity and verify the relevance of the VQA questions. Samples with ambiguous visuals or low-quality annotations are discarded. This rigorous human review results in the final set of 670 high-quality instances used in R$^3$-Bench.

\end{document}